\documentclass[11pt]{article}

\usepackage[preprint]{acl}

\usepackage{times}
\usepackage{latexsym}
\usepackage[T1]{fontenc}
\usepackage[utf8]{inputenc}
\usepackage{microtype}
\usepackage{inconsolata}

\usepackage[HTML,table,dvipsnames]{xcolor}
\usepackage{graphicx}
\usepackage{amsmath}
\usepackage{amssymb}
\usepackage{mathtools}
\usepackage{booktabs}
\usepackage{multirow}
\usepackage{enumitem}
\usepackage{subcaption}
\usepackage{fvextra}
\usepackage[most]{tcolorbox}
\usepackage{xspace}

\newcommand{\result}[2]{#1{\scriptsize$\pm$#2}}

\newcommand{\collabscirem}{\textsc{CollabSciReM}\xspace}
\newcommand{\collabllmsci}{\textsc{CollabLLMSci}\xspace}
\newcommand{\sciquality}{\textsc{SciQuality}\xspace}

\tcbset{promptbox/.style={
  colback=blue!5!white, colframe=blue!75!black,
  enhanced, breakable, fonttitle=\bfseries\small,
  left=2pt, right=2pt, top=2pt, bottom=2pt}}
\fvset{breaklines=true, breakanywhere=true, fontsize=\scriptsize,
  breaksymbolleft={}, breakindent=0pt}

\title{CollabSciReM: Aligning Language Models for Scientific Conversations via Science-Aware Reward Decomposition}

\author{
  Sasi Kiran Gaddipati\textsuperscript{1}\thanks{Corresponding author: \texttt{sasi.gaddipati@tib.eu}},
  Gollam Rabby\textsuperscript{2},
  S\"{o}ren Auer\textsuperscript{1,2} \\
  \textsuperscript{1}TIB—Leibniz Information Centre for Science and Technology, Hannover, Germany \\
  \textsuperscript{2}L3S Research Center, Hannover, Germany \\
}

\begin{document}
\maketitle

\begin{abstract}
Large Language Models (LLMs) are increasingly used as research collaborators for hypothesis refinement, methodological planning, and technical analysis across multi-turn interactions. In this setting, a useful assistant must be accurate, methodologically justified, logically consistent, and token-efficient. Current alignment approaches rarely achieve all of these together. Instruction-tuned models gain conversational quality through verbosity, and we show that science-agnostic general collaborative rewards compress responses at the expense of scientific rigor. We propose \textsc{CollabSciReM}, a goal-conditioned reward framework that decomposes the session-level reward into conversational quality, scientific rigor, and structural efficiency. We also introduce \textsc{SciQuality}, a pairwise evaluation of multi-turn research conversations grounded in 300 held-out scientific papers and scored on six dimensions: methodological soundness, justification, logical consistency, factual correctness, novelty, and responsiveness. Fine-tuned on 8.62K synthetic research conversations, \textsc{CollabSciReM} is preferred on \textsc{SciQuality} over a science-agnostic reward baseline trained on the same data for all four \textsc{Llama} and \textsc{Qwen} models, including a 65.3\% win rate at matched response length. For the 7B and 8B models, \collabscirem outperforms the science-agnostic reward baseline on the primary metric in 8 of 10 single- and multi-turn benchmark settings. In a blind study, domain experts rated \collabscirem significantly higher on novelty than the general-reward variant and reported fewer hallucinated responses than both the general-reward variant and the base model.
\end{abstract}

\section{Introduction}

\label{sec:introduction}

\begin{figure*}[t]
    \centering
    \includegraphics[width=\textwidth]{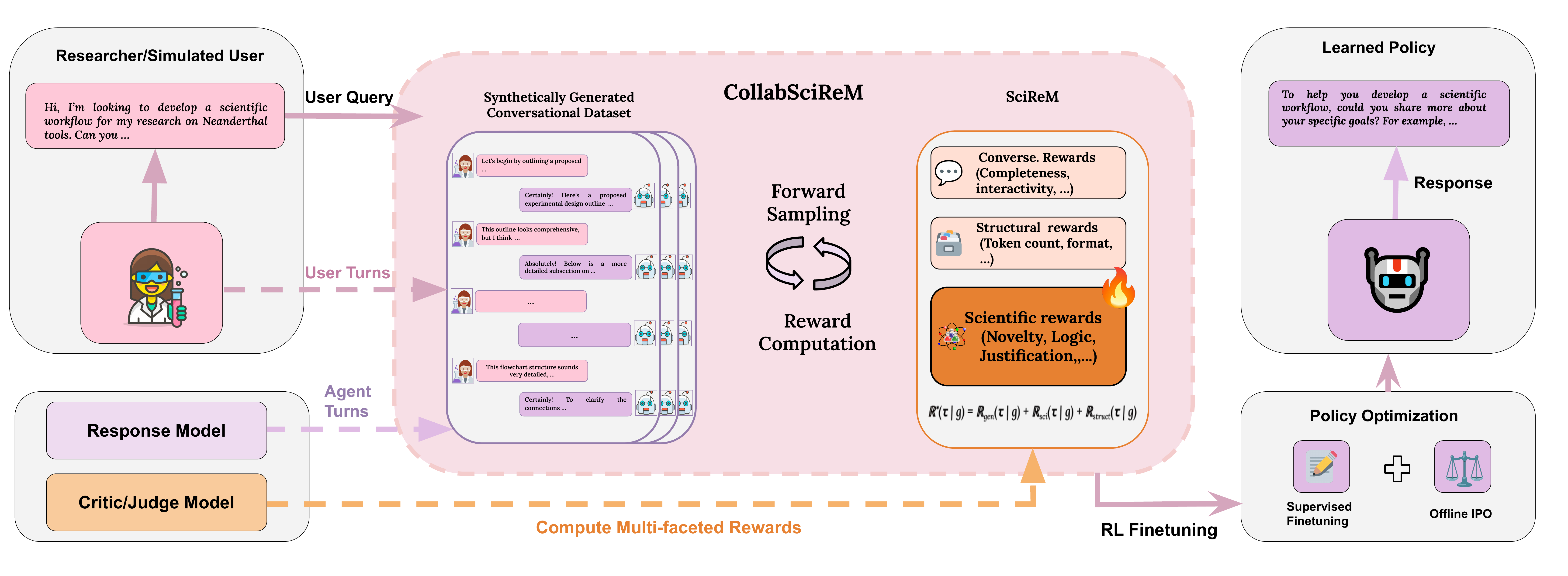}
    \caption{Overview of the \textsc{CollabSciReM} Framework. The pipeline begins with Synthetic Data Generation, represented by dotted arrows, in which a simulated researcher and a teacher model engage in scientific dialogue. These interactions are evaluated by a Critic/Judge Model to compute multi-faceted rewards, including conversational, structural, and scientific metrics. The resulting preference-labeled dataset is used to train a  Policy Model via Supervised Fine-Tuning (SFT) and Offline Identity Preference Optimization (IPO). In inference, a user query triggers the Learned Policy to generate a scientifically grounded response.}
    \label{fig:collab_scirem_framework}

\end{figure*}

Large Language Models (LLMs) are increasingly embedded in scientific research workflows, assisting with literature synthesis~\citep{lala2023paperqa, asai2024openscholar}, experimental planning~\citep{boiko2023autonomous, m2024augmenting}, hypothesis refinement~\citep{baek2025researchagent, rabby2024mc}, and scientific writing~\citep{gaddipati2025aissistant, lu2024ai}. Such collaboration typically unfolds over iterative, multi-turn conversations in which researchers refine objectives, clarify assumptions, revise methods, and scrutinize intermediate reasoning~\citep{kim2024understanding}. Effective scientific collaboration therefore depends on more than factual accuracy: recommendations must be methodologically sound, justifications valid and transparent, and reasoning consistent across the conversation.

Therefore, we believe that LLMs in scientific conversations must balance accuracy, scientific conversational quality, and structural efficiency, yet current alignment methods trade these properties against each other. Reinforcement learning from human feedback (RLHF) and related preference-learning methods~\citep{ouyang2022training, rafailov2023direct} optimize general helpfulness at the level of individual responses, producing long, elaborated outputs that LLM judges rate highly~\citep{dubois2024length, saito2023verbosity} but whose accuracy on expert-level reasoning remains low~\citep{rein2024gpqa, hendrycks2020measuring}. Session-level and multi-turn approaches~\citep{wu2025collabllm, mehri2026learning} extend optimization to interaction-level objectives for human--AI collaboration, but their rewards still target general conversational quality and task accuracy rather than the methodological and epistemic standards of scientific research. We show that training with such a general collaborative reward compresses responses and can improve accuracy, but the shorter responses deviate from soundness and valid justifications. As a result, aligned LLMs continue to exhibit hallucinations, weak methodological grounding, and unreliable reasoning in specialized domains~\citep{omar2025multi, agarwal2024medhalu, xiong2025toward}.

To this end, we introduce \textbf{Collab}orative \textbf{Sci}entific \textbf{Re}ward \textbf{M}odeling (\collabscirem) decomposing the session-level reward into three complementary reward functions, as shown in Figure~\ref{fig:collab_scirem_framework}. \textit{Conversational quality} measures how accurate, responsive, and understandable the interaction is. \textit{Scientific rigor} measures the methodological soundness of recommendations, the validity of justifications, and the logical consistency of reasoning. \textit{Structural efficiency} measures how well scientific information is organized and conveyed without unnecessary overhead. Our premise is that scientific conversational quality is multidimensional and cannot be captured by general preferences such as accuracy alone.

In addition, existing evaluations also leave a critical gap in assessing scientific research conversations. Scientific reasoning benchmarks primarily measure single-turn answer accuracy, while multi-turn benchmarks focus on general conversational quality. Neither evaluates whether research conversations are methodologically sound, well justified, and logically consistent across turns. To address this gap, we introduce \sciquality, a pairwise, LLM-as-a-judge evaluation framework for assessing the scientific quality of multi-turn research conversations grounded in held-out papers.

We fine-tune \textsc{Llama} and \textsc{Qwen} models using \collabscirem on synthetic research conversations and compare them against their instruction-tuned base models and \textsc{CollabLLMSci}, which uses the same training data but optimizes the science-agnostic general conversational \textsc{CollabLLM} reward~\citep{wu2025collabllm}. By holding the data, base models, and training recipe constant, we isolate the effect of reward design. Across all four model configurations, \collabscirem is preferred over \textsc{CollabLLMSci} on \sciquality, achieving a 65.3\% win rate even when response lengths are matched. A blind expert study finds significantly higher novelty than a general-reward variant and fewer hallucinated responses. Importantly, these gains do not compromise reasoning accuracy: \collabscirem improves expert-level reasoning accuracy over the base models by up to 9.09 percentage points while using 28--63\% fewer tokens, and outperforms \textsc{CollabLLMSci} in 8 of 10 single- and multi-turn benchmark settings.\footnote{Code, data, and the \sciquality evaluation are available at \url{https://github.com/gsasikiran/collabscirem}.}

Our contributions are as follows:

\begin{enumerate}
    \item \textsc{CollabSciReM}, a goal-conditioned, science-aware reward that decomposes collaborative reward into conversational quality, scientific rigor (justification, logical flow, novelty), and structural efficiency (format and token cost).
    \item We construct a training corpus of 8.62K multi-turn scientific conversations, sampled from 700 research question–methodology pairs spanning more than ten disciplines. We also introduce \sciquality, a pairwise evaluation of the scientific quality of multi-turn research conversations across six dimensions, built from 300 held-out scientific papers. We release the conversations, judge prompts, and scoring scripts.
    \item We show that \collabscirem balances accuracy, scientific conversational quality, and efficiency better than a general collaborative reward trained on the same data, across two model families, \textsc{Llama} and \textsc{Qwen}, supported by ablations of each reward component and a blind expert user study.
\end{enumerate}

\section{Related Work}

Recent works have proposed autonomous scientific agents to automate hypothesis generation, experimentation, and analysis \citep{lu2024ai, yamada2025ai, schmidgall2025agent, yu2025tinyscientist}. While these systems show promise, their reliability is constrained by the tendency of current LLMs to hallucinate or produce ungrounded reasoning \citep{omar2025multi, xiong2025toward, agarwal2024medhalu, gaddipati2025aissistant}. In fully autonomous settings, such errors can propagate into invalid scientific conclusions.  \citet{schmidgall2025agent} and \citet{gaddipati2025aissistant} highlight the effectiveness of human-in-the-loop designs, which substantially improve robustness and scientific validity. These findings motivate alignment frameworks that explicitly support sustained scientific human–AI collaboration rather than full autonomy.

Most LLM alignment methods, including RLHF and its variants, formulate learning as immediate turn-level optimization, assigning rewards to isolated responses \citep{schulman2017proximal, ouyang2022training, rafailov2023direct, lee2023rlaif}. This assumes that local response quality translates to successful interactions. However, LLM-assisted research collaborations are inherently sequential, requiring models to adapt to evolving hypotheses, partial evidence, and user intent over multiple turns. Consequently, immediate turn-level objectives may fail to capture how intermediate actions influence clarification trajectories and long-horizon task success \citep{zamfirescu2023johnny, wang2024understanding, kim2024understanding}.

Works by \citet{shani2024multi, zhou2024archer} and \citet{shi2024direct} have extended alignment to goal-conditioned multi-turn settings; these approaches typically assign rewards post-hoc over completed conversations rather than modeling the causal impact of intermediate decisions. While \citet{wu2025collabllm} and \citet{mehri2026learning} consider evolving dialogue states, their reward designs target general-purpose interaction quality and do not directly address the stringent correctness and grounding requirements in scientific contexts. In parallel, domain-specific autonomous works in mathematics  \citep{romera2024mathematical}, chemistry \citep{m2024augmenting, hayes2025simulating}, psychology \citep{wang2026perm}, and materials science \citep{merchant2023scaling, pyzer2022accelerating, zhang2024honeycomb} demonstrate strong performance within narrow scientific settings, but rely on specialized pipelines that limit generality across scientific domains. Other works \citep{auer2025towards, shao2025sciscigpt} focus on human-AI scientific workflows or iterative hypothesis refinement \citep{baek2025researchagent, li2024chain}, yet often underemphasize interaction-aware correctness, efficiency, and user intent evolution.
For evaluation, MT-Bench \citep{zheng2023judging} assesses general multi-turn conversational ability, and length-controlled evaluation \citep{dubois2024length} adjusts for the preference of LLM judges for longer responses. We introduce \sciquality{} to complement these benchmarks by targeting the scientific quality of research conversations. Together, these gaps highlight the need for alignment objectives that explicitly model scientific validity across domains, which \textsc{CollabSciReM} addresses by aligning intermediate model actions with long-horizon scientific objectives through science-aware reward decomposition grounded in scientific data distributions.

\section{Collaborative Scientific Reward Modeling}
 We propose \textsc{CollabSciReM}, built upon the work of \citet{wu2025collabllm}, leveraging the goal-conditioned session-level reward modeling with science-aware rewards, enabling better alignment with the demands of human–AI scientific conversations.

\subsection{Problem Formulation}

We consider a goal-conditioned scientific dialogue consisting of $K$ turns. Each turn $k$ is defined by the pair $(u_k, m_k)$, where $u_k$ represents the user query and $m_k$ denotes the model-generated response.

\subsubsection{State Representation}

At turn $j$, the model observes the dialogue history $\mathcal{H}$, which defines the current state $s_j = \mathcal{H}_j$. The state encapsulates all preceding interactions up to the current user input:
\begin{equation}
s_j = (u_1, m_1, u_2, m_2, \dots, u_j).
\end{equation}

\subsubsection{Policy and Trajectory}

Given the state $s_j$ and a specific goal $g$, the model generates a response $m_j$, its action at turn $j$, from a goal-conditioned policy $\pi_{\theta}$:
\begin{equation}
m_j \sim \pi_{\theta}(\cdot \mid s_j, g).
\end{equation}

A complete session trajectory $\tau$ is the sequence of all transitions over $K$ turns:
\begin{equation}
\tau = (t_1, t_2, \dots, t_K),
\end{equation}
where each transition is defined as
\begin{equation}
t_k = (u_k, m_k).
\end{equation}

\subsection{Goal-conditioned Multi-turn Aware Reward}

 \emph{Multi-turn Aware Reward} (MR)~\citep{wu2025collabllm} calculates the expected global utility of a response $m_j$ given the potential future trajectory $\tau_{j+1:K}$. This expected return of taking action $m_j$ in state $s_j$ is represented as:
\begin{equation}
\begin{aligned}
\mathcal{R}_{\text{MR}}(m_j \mid s_j, g)
&=
\mathbb{E}_{\tau_f \sim P(\tau_{j+1:K} \mid s_j, m_j)}
\\
&\quad \left[
\mathcal{R}^*(\tau \mid g)
\right]
\end{aligned}
\end{equation}
where $\tau_f$ represents the forward trajectory of the remaining conversation, and $\mathcal{R}^*$ is the holistic session-level reward function. However, computing the expected reward over all forward trajectories becomes exponentially expensive as the number of turns increases. To manage this, the turn depth window is limited to a maximum of 4 when calculating the expected conversation reward.

\subsection{Decomposing the Scientific Objective}

To ensure scientific and conversational efficiency, we decompose the session-level reward $\mathcal{R}^*(\tau \mid g)$ into three primary components: general conversational utility, scientific rigor, and structural efficiency. We formalize this as:
\begin{equation}
\label{eq:session_level_reward}
\begin{aligned}
\mathcal{R}^*(\tau \mid g)
&= \mathcal{R}_{\text{gen}}(\tau \mid g)
\\
&\quad + \mathcal{R}_{\text{sci}}(\tau \mid g)
+ \mathcal{R}_{\text{struct}}(\tau \mid g)
\end{aligned}
\end{equation}

\paragraph{General Conversational Utility ($\mathcal{R}_{\text{gen}}$)}

Existing goal-conditioned multi-turn rewards often focus on narrow metrics like generic accuracy and interaction. To capture the multidimensional nature of user interaction, we define $\mathcal{R}_{\text{gen}}$ as a weighted composite of $N_{\text{gen}}$ attributes:
\begin{equation}
\mathcal{R}_{\text{gen}}(\tau \mid g) = \sum_{i=1}^{N_{\text{gen}}} \gamma_i \, \phi_i(\tau \mid g)
\end{equation}
where
$\phi_i \in \{
\text{Accuracy},\,
\text{Interactivity},\,
\text{Clarity},\,$
\[
\text{Completeness}
\}.
\]

\paragraph{Scientific Rigor Metrics ($\mathcal{R}_{\text{sci}}$)}

We define $\mathcal{R}_{\text{sci}}$ as a weighted composite of $N_{\text{sci}}$ LLM-based evaluators:
\begin{equation}
\mathcal{R}_{\text{sci}}(\tau \mid g) = \sum_{i=1}^{N_{\text{sci}}} \gamma_i \, \phi_i(\tau \mid g)
\end{equation}
where $\phi_i \in \{\text{Justification, Logical Flow, Novelty}\}$ are scoring functions and $\gamma_i$ are their respective importance weights. This enables the reward model to penalize superficiality in favor of deep technical reasoning.

\paragraph{Efficiency and Structural Penalties ($\mathcal{R}_{\text{struct}}$)}

To mitigate \emph{verbosity sprawl}, a common failure mode in RLHF where models maximize reward through excessive length \citep{chen2024odin}, we extend the token-based penalty with a formatting reward metric. Therefore, the structural reward is defined as:
\begin{equation}
\mathcal{R}_{\text{struct}}(\tau \mid g) = \gamma_t (\lambda \cdot |\tau|_{\text{tokens}}) + \sum_{h=1}^{N_{\text{struct}}} \gamma_h \, \phi_h(\tau \mid g)
\end{equation}
where $|\tau|_{\text{tokens}}$ denotes the total token count, $\lambda$ is a scaling hyperparameter, and $\phi_h \in \{\text{Format}\}$. This encourages the model to maximize the information-to-token ratio, favoring concise scientific communication in required formats. We present and explain all the relevant parameters and scoring functions of the methodology in Appendix \ref{appendix: sci_metrics_and_weights}.

\section{Experimental setup}
\label{sec:experimentation}
\paragraph{Synthetic Data Generation}

We construct our fine-tuning conversational dataset from structured scientific articles, curated by \citet{laion2025ossas}, which organizes raw content into components such as research questions, methodological descriptions, experimental procedures, and model architectures. From this corpus, we sampled the initial 700 research question–methodology pairs and transformed these structured elements into 8.62K human--AI conversational sessions with multiple session samples generated for each pair. The distribution contains the data from diverse scientific disciplines, including Medicine (9.57\%), Computer Science (7.71\%), Public Health (5.57\%), Materials Science (4.86\%), Chemistry (4.71\%), Psychology (2.71\%), Education (2.71\%), Neuroscience (2.14\%), Oncology (2.00\%), Physics (2.00\%), and Others (56.02\%). In each conversation, user turns are grounded in research questions or clarification prompts, while assistant responses provide detailed descriptions of corresponding methodologies, experimental procedures, or architectural designs. We generate synthetic conversations using a strong instruction-following language model,  \emph{gpt-4o-mini} \citep{OpenAI2024GPT4omini}. Figure~\ref{fig:synthetic_data_excerpt} in Appendix~\ref{appendix:hyperparams} illustrates the pipeline from raw scientific papers to structured multi-turn conversations, along with the corresponding user and system prompts.

\paragraph{Reward Annotation}

We annotate each conversation with the previously defined multiple reward signals. We employ a critic model, \emph{gpt-5-mini} \citep{singh2025openai} to evaluate and score reward metrics for each turn given the future trajectories of the conversation. We adopt LLM-based reward scoring as many desired properties, such as scientific logic, completeness, and novelty, are difficult to capture using rule-based evaluators, and they enable scalable and consistent evaluation across large datasets at significantly lower cost than expert human annotation.

\paragraph{Fine-Tuning Base Models} To assess whether the reward modeling strategy generalizes across architectures, we fine-tune models from both the Llama~\citep{grattafiori2024llama} and Qwen~\citep{yang2024qwen2} families on our synthetically generated scientific conversational dataset. Fine-tuning proceeds in two stages: supervised fine-tuning (SFT), followed by offline preference optimization using the Identity Preference Optimization (IPO) objective~\citep{azar2024general} on both \textsc{CollabLLM} and \textsc{CollabSciReM} rewards. Negative sampling is performed by rejecting the conversation samples that receive lower scores, according to the selected reward model. All experiments were conducted on 4$\times$ NVIDIA H100 GPUs (80 GB). Appendix~\ref{appendix:hyperparams} provides the relevant hyperparameters in detail.

\begin{figure}[t]
    \centering
    \includegraphics[width=0.75\columnwidth]{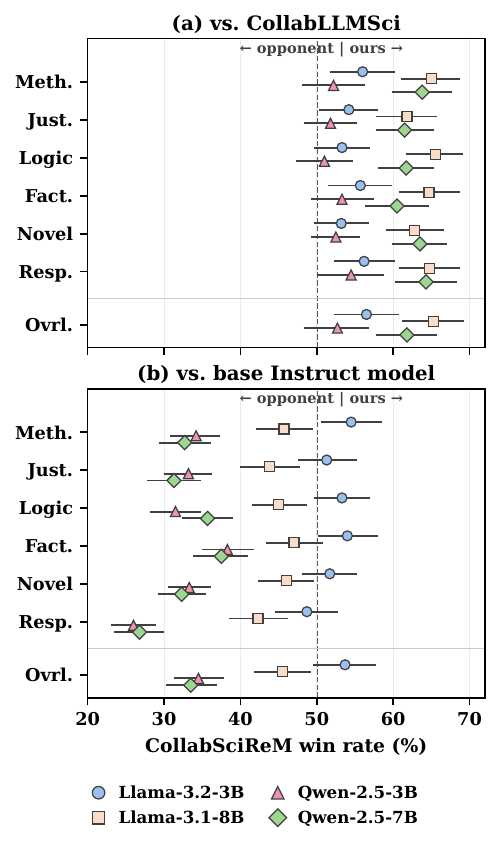}
    \caption{Pairwise scientific-quality win rates of CollabSciReM against (a) CollabLLMSci and (b) the base Instruct models on 300 held-out scientific papers. Bars show 95\% confidence intervals. \textbf{Meth.}: Methodological soundness, \textbf{Just.}: Justification, \textbf{Logic}: Logical Consistency, \textbf{Fact.}: Factual Correctness, \textbf{Novel}: Novelty, \textbf{Resp.}: Responsiveness and \textbf{Ovrl.}: Overall}
    \label{fig:sciquality_main}
\end{figure}

\begin{table*}[t]
\centering
\caption{Performance of 7B and 8B models on (a) multi-turn benchmarks (MT-Bench-STEM, MINT) and (b) single-turn benchmarks (GPQA-D, MMLU-Valid, TheoremQA). \textbf{Acc}: accuracy (\%). \textbf{SR@5}: success rate within five interaction turns (\%). \textbf{TC}: average tokens per response (per turn for MT-Bench-STEM). \textbf{APT}: accuracy per token, computed as $\text{Acc} / \text{TC} \times 1000$ (SR@5 in place of Acc for MINT). Results are mean $\pm$ sample standard deviation over three seeds; bold values indicate the best performance in each column within each model family.}
\label{tab:evaluation_results}
\small
\setlength{\tabcolsep}{3.5pt}
\resizebox{\textwidth}{!}{%
\begin{tabular}{@{}l *{9}{c}@{}}
\toprule
\multicolumn{10}{l}{\textit{(a) Multi-turn benchmarks}} \\
\midrule
& \multicolumn{6}{c}{\textbf{MT-Bench-STEM}} & \multicolumn{3}{c}{\textbf{MINT}} \\
\cmidrule(lr){2-7} \cmidrule(lr){8-10}
\textbf{Model} & \textbf{Overall} ($\uparrow$) & \textbf{Acc.} ($\uparrow$) & \textbf{Format} ($\uparrow$) & \textbf{Useful.} ($\uparrow$) & \textbf{Relev.} ($\uparrow$) & \textbf{TC} ($\downarrow$) & \textbf{SR@5 (\%)} ($\uparrow$) & \textbf{TC} ($\downarrow$) & \textbf{APT} ($\uparrow$) \\
\midrule
Llama-3.1-8B-Instruct &
\result{6.83}{0.04} & \result{6.55}{0.05} & \result{7.92}{0.07} & \result{6.67}{0.08} & \result{8.37}{0.01} & \result{469.00}{8.00} &
\textbf{\result{47.11}{3.20}} & \result{296.00}{14.06} & \result{159.20}{9.08} \\
Llama-3.1-8B-CollabLLMSci &
\result{7.02}{0.19} & \result{6.69}{0.22} & \textbf{\result{8.19}{0.01}} & \result{6.88}{0.22} & \textbf{\result{8.55}{0.02}} & \textbf{\result{298.00}{6.00}} &
\result{43.41}{3.22} & \textbf{\result{123.91}{0.62}} & \textbf{\result{350.25}{24.14}} \\
Llama-3.1-8B-CollabSciReM &
\textbf{\result{7.15}{0.20}} & \textbf{\result{6.84}{0.25}} & \textbf{\result{8.19}{0.04}} & \textbf{\result{6.94}{0.16}} & \result{8.52}{0.08} & \result{310.00}{4.00} &
\result{36.86}{0.89} & \result{147.60}{8.63} & \result{250.47}{15.27} \\
\midrule
Qwen-2.5-7B-Instruct &
\textbf{\result{7.91}{0.09}} & \result{7.62}{0.11} & \textbf{\result{8.70}{0.07}} & \textbf{\result{7.80}{0.04}} & \result{8.90}{0.03} & \result{555.00}{10.00} &
\result{66.07}{1.03} & \result{226.34}{16.16} & \result{292.73}{16.98} \\
Qwen-2.5-7B-CollabLLMSci &
\result{7.67}{0.05} & \result{7.40}{0.14} & \textbf{\result{8.70}{0.11}} & \result{7.48}{0.04} & \result{8.85}{0.09} & \result{418.00}{5.00} &
\result{68.59}{1.28} & \textbf{\result{112.15}{2.62}} & \textbf{\result{611.95}{23.60}} \\
Qwen-2.5-7B-CollabSciReM &
\result{7.85}{0.11} & \textbf{\result{7.64}{0.15}} & \result{8.68}{0.05} & \result{7.63}{0.21} & \textbf{\result{8.93}{0.16}} & \textbf{\result{399.00}{4.00}} &
\textbf{\result{69.63}{1.12}} & \result{120.54}{2.01} & \result{577.84}{18.31} \\
\midrule
\multicolumn{10}{l}{\textit{(b) Single-turn benchmarks}} \\
\midrule
& \multicolumn{3}{c}{\textbf{GPQA-D}} & \multicolumn{3}{c}{\textbf{MMLU-Valid}} & \multicolumn{3}{c}{\textbf{TheoremQA}} \\
\cmidrule(lr){2-4} \cmidrule(lr){5-7} \cmidrule(lr){8-10}
\textbf{Model} & \textbf{Acc (\%)} ($\uparrow$) & \textbf{TC} ($\downarrow$) & \textbf{APT} ($\uparrow$) & \textbf{Acc (\%)} ($\uparrow$) & \textbf{TC} ($\downarrow$) & \textbf{APT} ($\uparrow$) & \textbf{Acc (\%)} ($\uparrow$) & \textbf{TC} ($\downarrow$) & \textbf{APT} ($\uparrow$) \\
\midrule
Llama-3.1-8B-Instruct &
\result{22.56}{1.25} & \result{1644.00}{47.74} & \result{13.71}{0.62} &
\result{48.97}{0.81} & \result{702.79}{2.37} & \result{69.68}{1.38} &
\textbf{\result{19.28}{0.50}} & \result{1351.30}{46.23} & \result{14.27}{0.21} \\
Llama-3.1-8B-CollabLLMSci &
\result{29.97}{1.91} & \textbf{\result{478.71}{1.67}} & \textbf{\result{62.60}{3.99}} &
\result{57.26}{1.14} & \textbf{\result{380.88}{1.12}} & \textbf{\result{150.34}{2.96}} &
\result{16.73}{0.67} & \textbf{\result{396.72}{13.39}} & \textbf{\result{42.21}{2.16}} \\
Llama-3.1-8B-CollabSciReM &
\textbf{\result{31.65}{1.04}} & \result{611.40}{29.74} & \result{51.97}{4.21} &
\textbf{\result{57.63}{0.66}} & \result{403.68}{3.97} & \result{142.77}{1.38} &
\result{17.49}{0.17} & \result{505.70}{18.60} & \result{34.63}{1.16} \\
\midrule
Qwen-2.5-7B-Instruct &
\result{29.46}{2.27} & \result{883.25}{7.11} & \result{33.35}{2.51} &
\textbf{\result{65.99}{0.51}} & \result{479.62}{1.17} & \result{137.59}{1.22} &
\textbf{\result{27.80}{0.41}} & \result{603.57}{6.46} & \result{46.07}{1.17} \\
Qwen-2.5-7B-CollabLLMSci &
\result{28.96}{1.91} & \result{488.24}{3.92} & \result{59.29}{3.46} &
\result{64.53}{0.17} & \result{335.22}{0.50} & \result{192.51}{0.32} &
\result{24.63}{0.75} & \textbf{\result{372.20}{0.80}} & \textbf{\result{66.18}{2.14}} \\
Qwen-2.5-7B-CollabSciReM &
\textbf{\result{33.50}{2.74}} & \textbf{\result{482.30}{10.14}} & \textbf{\result{69.58}{6.88}} &
\result{64.27}{0.75} & \textbf{\result{316.69}{0.34}} & \textbf{\result{202.95}{2.20}} &
\result{25.12}{0.25} & \result{380.30}{1.06} & \result{66.06}{0.68} \\
\bottomrule
\end{tabular}%
}
\end{table*}

\paragraph{\sciquality} We select 300 papers from the OSSAS corpus~\citep{laion2025ossas} that are disjoint from the 700 used to generate the training data. For each paper, a fine-tuned model acts as a user simulator, seeded with the paper's research question, holds a four-turn conversation with each model. \textsc{DeepSeek-V4-Pro} then compares the conversations of two models pairwise. We evaluate each pair in both orders to control for position bias and report win rates with 95\% bootstrap confidence intervals. We choose \textsc{DeepSeek} as the judge because it belongs to a different model family than our training critic (\textsc{gpt-5-mini}), which reduces the risk that the evaluation simply rewards the critic's own preferences. The judge prompt and rubric are given in Appendix~\ref{app:sciq-judge}.

\section{Benchmark Evaluation}
\label{sec:benchmark_eval}

We evaluate our hypothesis on three multi-turn benchmarks, \sciquality,  MT-Bench-STEM~\citep{zheng2023judging} and MINT~\citep{wang2024mint}, and on three single-turn scientific reasoning benchmarks: GPQA-D~\citep{rein2024gpqa}, MMLU-Validation~\citep{hendrycks2020measuring}, and TheoremQA~\citep{chen2023theoremqa}. For \textsc{Llama-3.1-8B-Instruct} and \textsc{Qwen-2.5-7B-Instruct}, we compare the base model, \collabllmsci, and \collabscirem.\footnote{The fine-tuned \textsc{Llama-3.1-8B} model is available at \url{https://huggingface.co/gsasikiran/collabscirem-Llama-3.1-8B-Instruct}.} The two fine-tuned models differ only in their reward, so comparing them isolates the effect of science-aware reward modeling.

We report pass@1 accuracy on the single-turn benchmarks, excluding image-based instances, and the success rate within five turns, SR@5 on MINT.  \emph{MT-Bench-STEM} consists of the 20 two-turn questions from the STEM and Math categories of MT-Bench, and \texttt{deepseek-v4-pro-0813}~\citep{deepseek2026v4} scores each response from 1 to 10 on overall quality, accuracy, format, usefulness, and relevance. We measure efficiency with the average number of tokens per response, TC, and accuracy per token, $\text{APT}=\text{Acc}/\text{TC}\times 1000$, using SR@5 in place of accuracy for MINT. All models are evaluated zero-shot with free generation, using the same prompt, decoding parameters, 4,096-token generation budget, and answer-extraction function. Table~\ref{tab:evaluation_results} illustrates the results, averaged over three seeds.

\begin{table*}[t]
\centering
\caption{Ablation and sensitivity analysis of reward components for \textsc{CollabSciReM} on Llama-3.1-8B-Instruct, fine-tuned with SFT followed by offline IPO. The upper block reports multi-turn benchmarks, MT-Bench-STEM and MINT, and the lower block reports single-turn benchmarks, GPQA-D, MMLU-Valid, and TheoremQA. $R_{\text{gen}}$, $R_{\text{struct}}$, and $R_{\text{sci}}$ denote the conversational, structural, and science-aware rewards, respectively.}
\label{tab:ablation_results}
\small
\setlength{\tabcolsep}{3.5pt}
\resizebox{\textwidth}{!}{%
\begin{tabular}{@{}l *{9}{c}@{}}
\toprule
\multicolumn{10}{l}{\textit{(a) Multi-turn benchmarks}} \\
\midrule
& \multicolumn{6}{c}{\textbf{MT-Bench-STEM}} & \multicolumn{3}{c}{\textbf{MINT}} \\
\cmidrule(lr){2-7} \cmidrule(lr){8-10}
\textbf{Reward Model} & \textbf{Overall} ($\uparrow$) & \textbf{Acc.} ($\uparrow$) & \textbf{Format} ($\uparrow$) & \textbf{Useful.} ($\uparrow$) & \textbf{Relev.} ($\uparrow$) & \textbf{TC} ($\downarrow$) & \textbf{SR@5 (\%)} ($\uparrow$) & \textbf{TC} ($\downarrow$) & \textbf{APT} ($\uparrow$) \\
\midrule
$R_{\text{gen}} + R_{\text{struct}}$ &
\result{6.81}{0.23} & \result{6.48}{0.23} & \result{8.03}{0.09} & \result{6.67}{0.26} & \result{8.38}{0.07} & \result{319}{3.00} &
\result{40.59}{2.57} & \textbf{\result{139.31}{1.87}} & \textbf{\result{291.45}{19.60}} \\
$R_{\text{gen}} + R_{\text{sci}}$ &
\result{6.58}{0.26} & \result{6.33}{0.25} & \result{7.74}{0.27} & \result{6.46}{0.43} & \result{8.10}{0.26} & \result{622}{15.00} &
\textbf{\result{48.30}{4.45}} & \result{699.95}{57.14} & \result{69.05}{4.17} \\
$R_{\text{struct}} + R_{\text{sci}}$ &
\result{6.92}{0.15} & \result{6.67}{0.10} & \result{8.06}{0.19} & \result{6.67}{0.12} & \result{8.30}{0.17} & \result{319}{8.00} &
\result{40.74}{2.72} & \result{149.81}{0.99} & \result{271.88}{16.36} \\
\midrule
\textsc{CollabSciReM} $\mid 0.7\,\gamma_{\textsc{CSciReM}}$ &
\textbf{\result{7.14}{0.14}} & \textbf{\result{6.84}{0.08}} & \textbf{\result{8.27}{0.09}} & \textbf{\result{6.85}{0.15}} & \textbf{\result{8.52}{0.23}} & \textbf{\result{308}{6.00}} &
\result{41.78}{4.00} & \result{148.24}{6.28} & \result{282.92}{39.05} \\
\midrule
\multicolumn{10}{l}{\textit{(b) Single-turn benchmarks}} \\
\midrule
& \multicolumn{3}{c}{\textbf{GPQA-D}} & \multicolumn{3}{c}{\textbf{MMLU-Valid}} & \multicolumn{3}{c}{\textbf{TheoremQA}} \\
\cmidrule(lr){2-4} \cmidrule(lr){5-7} \cmidrule(lr){8-10}
\textbf{Reward Model} & \textbf{Acc (\%)} ($\uparrow$) & \textbf{TC} ($\downarrow$) & \textbf{APT} ($\uparrow$) & \textbf{Acc (\%)} ($\uparrow$) & \textbf{TC} ($\downarrow$) & \textbf{APT} ($\uparrow$) & \textbf{Acc (\%)} ($\uparrow$) & \textbf{TC} ($\downarrow$) & \textbf{APT} ($\uparrow$) \\
\midrule
$R_{\text{gen}} + R_{\text{struct}}$ &
\result{29.63}{2.92} & \result{552.46}{27.71} & \result{53.59}{3.86} &
\result{57.00}{0.59} & \result{397.27}{2.97} & \result{143.49}{2.51} &
\result{15.93}{0.81} & \result{515.09}{30.40} & \result{31.06}{3.29} \\
$R_{\text{gen}} + R_{\text{sci}}$ &
\result{23.06}{0.77} & \result{1966.21}{89.88} & \result{11.75}{0.84} &
\result{42.24}{0.69} & \result{947.54}{20.01} & \result{44.60}{1.63} &
\textbf{\result{18.52}{0.69}} & \result{1344.53}{46.05} & \result{13.79}{0.82} \\
$R_{\text{struct}} + R_{\text{sci}}$ &
\result{31.14}{5.11} & \textbf{\result{510.67}{8.22}} & \textbf{\result{60.93}{9.51}} &
\textbf{\result{58.35}{0.87}} & \result{397.60}{3.46} & \textbf{\result{146.77}{3.13}} &
\result{15.62}{1.67} & \result{505.20}{17.07} & \result{30.98}{3.96} \\
\midrule
\collabllmsci $\mid \gamma_{\textsc{CSciReM}}$ &
\result{22.73}{1.24} & \result{1295.48}{74.12} & \result{17.64}{1.88} &
\result{52.36}{1.37} & \result{591.35}{4.80} & \result{87.04}{0.70} &
\result{17.47}{0.47} & \result{1063.15}{23.75} & \result{16.44}{0.08} \\
\textsc{CollabSciReM} $\mid 0.7\,\gamma_{\textsc{CSciReM}}$ &
\textbf{\result{31.31}{2.08}} & \result{580.33}{30.94} & \result{54.76}{5.74} &
\result{58.22}{0.29} & \textbf{\result{396.85}{7.29}} & \result{146.74}{3.17} &
\result{16.73}{1.01} & \textbf{\result{391.33}{19.06}} & \textbf{\result{42.88}{4.19}} \\
\bottomrule
\end{tabular}%
}
\end{table*}

\paragraph{Multi-turn interaction.}
On MT-Bench-STEM, \textsc{Llama-CollabSciReM} achieves the best overall score of 7.15, ahead of 6.83 for the base model and 7.02 for \textsc{CollabLLMSci}. It also scores highest on accuracy and usefulness while using 33.9\% fewer tokens than the base model. For \textsc{Qwen}, its overall score of 7.85 is within one standard deviation of the base model's 7.91. It leads its family on accuracy with 7.64 and on relevance with 8.93, and it produces the shortest responses at 399 tokens per turn, 28.1\% fewer than the base model.

On MINT, \textsc{Qwen-CollabSciReM} reaches the highest SR@5 of 69.63\%, compared with 66.07\% for the base model and 68.59\% for \textsc{CollabLLMSci}, while halving TC and roughly doubling APT. With \textsc{Llama}, both fine-tuned models fall below the base model, and \textsc{CollabSciReM} drops from 47.11\% to 36.86\%.

\paragraph{Scientific conversational quality} Figure~\ref{fig:sciquality_main} reports \sciquality win rates. Against \collabllmsci, \collabscirem is preferred on overall quality for all four models. For \textsc{Llama-3.1-8B}, both models produce 474 tokens per turn, so the 65.3\% preference cannot be attributed to response length. Against the base models, \collabscirem is at parity for \textsc{Llama-3.2-3B} and less preferred for the other three models (33.5--45.5\%). The base Instruct models produce responses 1.6--3.1$\times$ longer than their fine-tuned counterparts. Because LLM judges tend to favor longer outputs, this comparison likely reflects response length as well as quality, consistent with the user study in Section~\ref{sec:human_eval}.

\paragraph{Single-turn scientific reasoning.}
Gains are largest on GPQA-D, which requires multi-step, expert-level reasoning. \textsc{Llama-CollabSciReM} raises accuracy from 22.56\% to 31.65\%, a gain of 9.09 points that also exceeds the 29.97\% of \textsc{CollabLLMSci}. It does so while cutting tokens by 62.8\%, which increases APT 3.8-fold. For \textsc{Qwen}, it is the only fine-tuned variant to surpass the base model, achieving 33.50\% against 29.46\% for the base model and 28.96\% for \textsc{CollabLLMSci}. It also produces the shortest responses in its family and reaches 2.1 times the base model's APT. On MMLU-Validation, \textsc{Llama-CollabSciReM} improves accuracy from 48.97\% to 57.63\%, a gain of 8.66 points, with 42.6\% fewer tokens, doubling APT. The \textsc{Qwen} base model is already strong on this benchmark, and \textsc{CollabSciReM} gives up 1.72 points of accuracy in exchange for a 34.0\% token reduction and an APT 47.5\% higher than the base model's, the best in its family. TheoremQA is the exception, as neither fine-tuned model matches the base model in either family, reflecting the mathematical data distribution difference between fine-tuned and test datasets. \textsc{Llama-CollabSciReM} reaches 17.49\% against 19.28\% for the base model, and \textsc{Qwen-CollabSciReM} reaches 25.12\% against 27.80\%. Even so, \textsc{CollabSciReM} still outperforms \textsc{CollabLLMSci} in both families and uses 37 to 63\% fewer tokens than the base models.
Taken together, the three systems occupy distinct points in the trade-off between accuracy, scientific conversational quality, and efficiency. The base models produce the longest responses and are preferred by the judge on \sciquality, but they have the lowest accuracy on expert-level reasoning. \collabllmsci compresses responses substantially, but its accuracy gains are inconsistent across families. It improves over the base model on GPQA-D and MMLU-Validation for \textsc{Llama} but falls below it on both for \textsc{Qwen}, and it loses to \collabscirem on \sciquality for all four models. \collabscirem matches the compression of \collabllmsci while retaining scientific conversational quality. It achieves the highest GPQA-D accuracy in both families and scores higher than \collabllmsci on the primary metric in 8 of 10 benchmark–family settings. The two exceptions are \textsc{Llama} on \textsc{MINT} and \textsc{Qwen} on \textsc{MMLU-Valid}. On TheoremQA, both fine-tuned models trail the base model. This pattern supports our central claim: a general collaborative reward learns to be concise, while a science-aware reward learns to be concise without discarding scientific substance. We extend our evaluation on 3B models and present them in Appendix \ref{appendix: 3b_results}.

\begin{figure*}[t]
    \centering
    \begin{subfigure}{0.3\textwidth}
        \includegraphics[width=\linewidth]{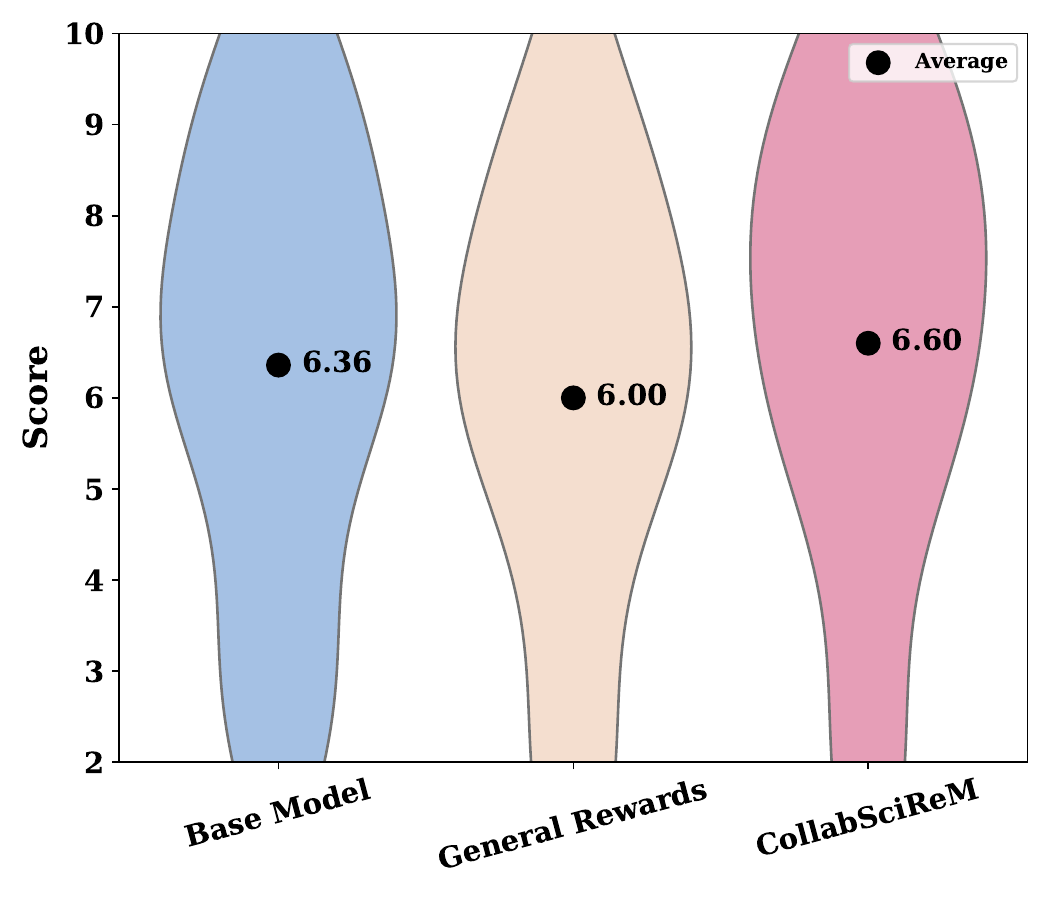}
        \caption{Logical Justification ($\uparrow$)}
        \label{fig:logical}
    \end{subfigure}
    \hfill
    \begin{subfigure}{0.3\textwidth}
        \includegraphics[width=\linewidth]{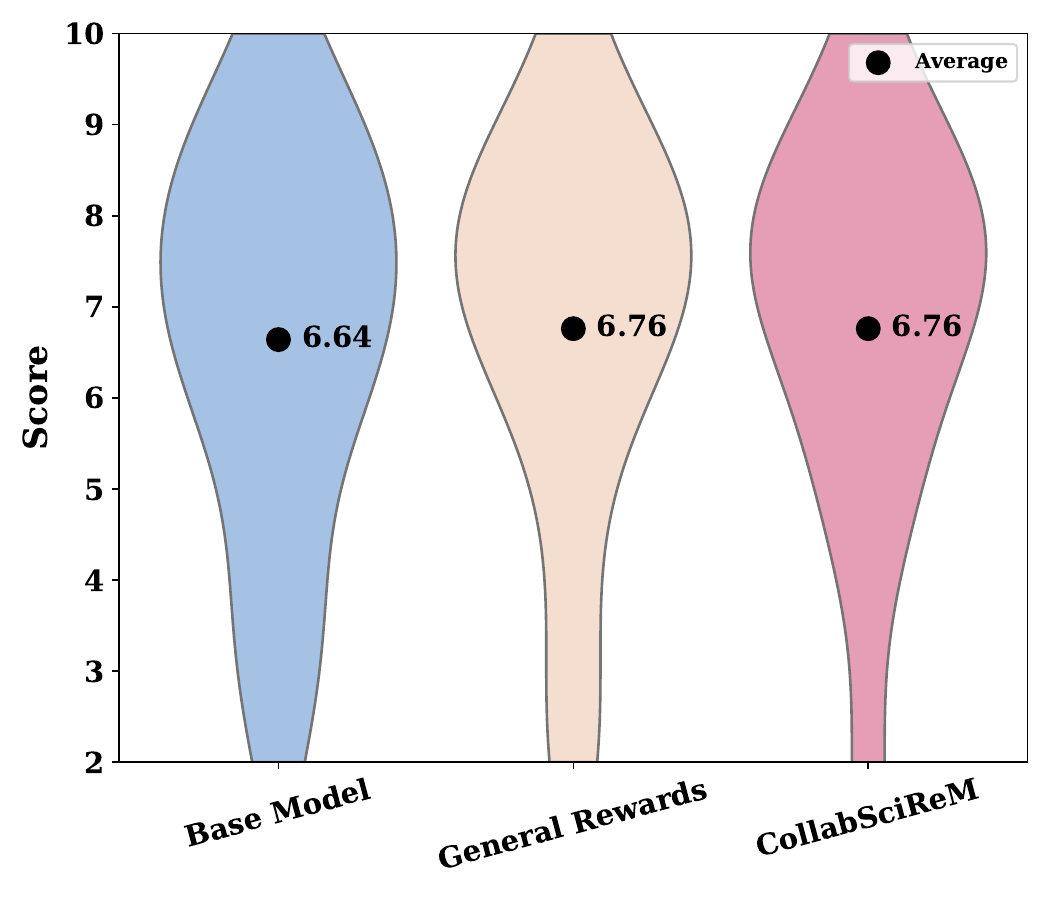}
        \caption{Interaction Quality ($\uparrow$)}
        \label{fig:interactive}
    \end{subfigure}
    \hfill
    \begin{subfigure}{0.3\textwidth}
        \includegraphics[width=\linewidth]{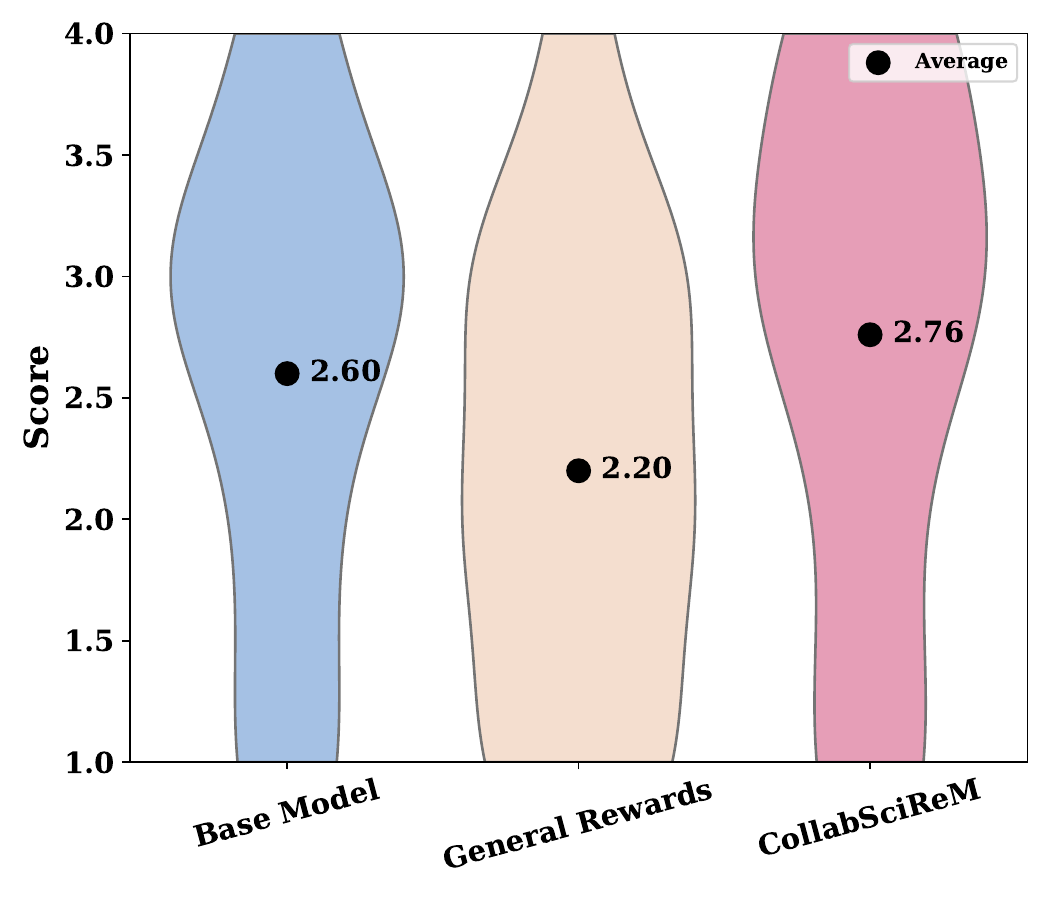}
        \caption{Novelty ($\uparrow$)}
        \label{fig:novelty}
    \end{subfigure}
    \hfill

    \caption{Real-world user study results showcasing participant interactions with anonymized base \textsc{Llama-3.1-8B-Instruct}, general rewards (accuracy and interaction, w/o token penalty) and \textsc{CollabSciReM} models. Users rated (a) logical justification (1-10), (b) overall interactive experience (1-10), and (c) novelty (1-5). }
    \label{fig:human_evaluation_results}
\end{figure*}
\section{Ablation Study}
\label{sec:ablation_study}

We ablate the \textsc{CollabSciReM} reward on \textsc{Llama-3.1-8B-Instruct} by removing one component at a time: the conversational reward $R_{\text{gen}}$, the structural reward $R_{\text{struct}}$, or the science-aware reward $R_{\text{sci}}$. We also train \textsc{CollabLLM} with the \textsc{CollabSciReM} weights and test sensitivity by scaling all weights by 0.7 and report the relevant results in Table~\ref{tab:ablation_results}.

Removing $R_{\text{sci}}$ lowers accuracy at a nearly unchanged token budget. MT-Bench-STEM overall score falls from 7.15 to 6.81, GPQA-D accuracy from 31.65\% to 29.63\%, TheoremQA from 17.49\% to 15.93\%. This shows that the science-aware reward makes reasoning more accurate rather than longer. Removing $R_{\text{struct}}$ inflates responses to 1966 tokens on GPQA-D and drops accuracy to 23.06\% on GPQA-D and 42.24\% on MMLU-Validation. The extended reasoning helps only computation-heavy tasks such as TheoremQA and MINT. Removing $R_{\text{gen}}$ preserves single-turn accuracy but lowers MT-Bench-STEM to 6.92 overall, with usefulness falling from 6.94 to 6.67, indicating that the conversational reward sustains interaction quality.

\textsc{CollabLLM} trained with the \textsc{CollabSciReM} weights reaches only 22.73\% on GPQA-D, matching the base model, with responses 1.5--2.1$\times$ longer than \collabscirem. The gains therefore stem from the science-aware reward signal rather than its weighting. Scaling all weights by 0.7 leaves most results nearly unchanged, with 31.31\% on GPQA-D and 7.14 overall on MT-Bench-STEM; the largest shift is on MINT, where SR@5 rises from 36.86\% to 41.78\%. This indicates robustness to reward scale. This finding also provides the flexibility to control the relevant rewards in scientific contexts as required. Together, $R_{\text{sci}}$ drives scientific correctness, $R_{\text{struct}}$ keeps reasoning concise, and $R_{\text{gen}}$ maintains conversational quality.

\section{User study}
\label{sec:human_eval}

To assess real-world utility, we conducted a blind user study on Prolific with subject matter experts holding advanced degrees in fields spanning computer science, manufacturing, media, medicine, law, and the social sciences. Each participant held a multi-turn scientific dialogue of at least three turns with each of three anonymized models: the base \textsc{Llama-3.1-8B-Instruct}, a variant fine-tuned with only general conversational rewards (\textit{accuracy and interaction}), and \textsc{CollabSciReM}. Participants then rated each model on logical justification (1--10), interaction quality (1--10), and novelty (1--5). The novelty uses a 1–5 scale since assessing originality is inherently coarser, and a larger scale would introduce unnecessary rater noise without increasing discriminative power. Recruitment criteria, interface design, and prompt guidelines are given in Appendix~\ref{appendix: user_study_setup}, and complete interaction logs are included in the supplementary material. \textsc{CollabSciReM} achieves the highest mean scores for logical justification at 6.6 and novelty at 2.76, and its interaction quality score of 6.76 exceeds the base model and matches the general-reward variant. Within-participant Wilcoxon signed-rank tests with 10,000-sample bootstrap confidence intervals show that \textsc{CollabSciReM} significantly outperforms the general-reward variant on novelty, with a mean difference of 0.56, a 95\% confidence interval from 0.16 to 0.96, $p = 0.023$, and a rank-biserial effect size of 0.758. Logical justification shows a positive but non-significant trend at $p = 0.085$. Participants also rated 52\% of \textsc{CollabSciReM} responses as free of hallucinations, compared with 44\% for the general-reward variant and 40\% for the base model.

\section{Conclusion}
\label{sec:conclusion}
We introduced \collabscirem, a science-aware reward that decomposes collaborative reward into conversational quality, scientific rigor, and structural efficiency, together with \sciquality, an evaluation of the scientific quality of multi-turn research conversations. Where general collaborative rewards, \collabllmsci, favor conciseness and instruction-tuned models favor elaboration, \collabscirem balances accuracy, scientific conversational quality, and efficiency. Ablations show that each reward dimension plays a distinct role, and a study with domain experts confirms a significant gain in novelty and fewer hallucinated responses.

\section*{Limitations}
\label{sec:limitations}

Several limitations remain. The gains of \collabscirem over a general collaborative reward are not uniform: it trails \collabllmsci on MINT with \textsc{Llama-3.1-8B} and on several 3B settings, and several benchmark differences fall within one standard deviation across seeds. First, the reward gains are strongest on data close to the training distribution. On out-of-distribution tasks such as the mathematical problems in TheoremQA and MINT, the base models remain competitive, and broadening the training corpus toward these domains is a natural next step. Second, our experiments use offline Identity Preference Optimization (IPO) to fit our computational budget. Online reinforcement learning methods such as PPO or GRPO offer a promising path toward stronger generalization. Third, our pipeline uses \texttt{gpt-4o-mini} for user simulation and \texttt{gpt-5-mini} for reward annotation. A systematic study of how teacher and critic strength shape downstream performance is an important avenue for future research. Finally, our literature-derived synthetic dialogues provide a scalable training signal, and incorporating real human-expert interaction data would further improve the fidelity and ecological validity of the preference signals. Collecting such data at scale requires substantial cost and coordination with subject matter experts, which makes it a valuable target for future collaborative work.

\section*{Ethical Considerations}

This work involves the development and evaluation of large language models (LLMs) for assisting scientific research through interactive, goal-conditioned dialogues. We carefully considered ethical implications across data generation, model finetuning, human evaluation, and potential deployment. Subject matter experts were recruited through a crowdsourcing platform, \textit{Prolific}, and compensated at an average rate of \$22.19 per hour, which exceeded the local minimum wage, with a median completion time of approximately 29.5 minutes. Data collected from participants was anonymized and used solely for research purposes. No personally identifiable information was retained beyond what was necessary for compensation processing, and this was clearly communicated to participants during the evaluation.

Despite our emphasis on reducing hallucinations and improving methodological rigor, LLMs remain susceptible to errors. Over-reliance on LLM-generated scientific content could propagate subtle inaccuracies or biases, particularly in high-stakes domains such as medicine, chemistry, or defense research. We strongly recommend that outputs from these LLMs be treated as collaborative aids rather than authoritative sources, with final responsibility resting with human researchers. Our framework’s focus on justification, logical flow, and novelty aims to make such limitations more transparent to users. Also, we recognize that advanced AI tools may widen gaps between well-resourced institutions and others. Therefore, we release our code, synthetic datasets, and models to promote transparency, reproducibility, and equitable access to this research.

We use \textsc{Llama-3.1-8B-Instruct} and \textsc{Qwen-2.5} models under their respective open licenses (Meta Llama 3.1 Community License and Qwen license). \textsc{gpt-4o-mini} and \textsc{gpt-5-mini} were accessed via OpenAI’s API for research purposes only, in accordance with their terms of service. The synthetic data generation is based on the OSSAS corpus \cite{laion2025ossas}. We release our synthetic dataset, code, and fine-tuned models under a permissive open-source license, consistent with the research-oriented nature of the source materials and compatible with the original licenses. We used Claude Code to assist with coding tasks and Claude Opus 5.5 to assist with rephrasing and summarizing text. The authors take full responsibility for all claims, results, and citations presented in this paper.

\section*{Acknowledgements}

This research was supported by the Federal Ministry of Education and Research (BMBF) under grant no. 01IS22093C Kisski, project 460234259 NFDI4DataScience consortium and project "WestAI0096", Jureca Computation from Jülich Supercomputing Centre (JSC). We thank Lokesh Veeramacheneni for his suggestions throughout the research and revisions of the manuscript.

\bibliography{references}

@article{agarwal2024medhalu,
  title={{MedHalu}: Hallucinations in Responses to Healthcare Queries by Large Language Models},
  author={Agarwal, Vibhor and Jin, Yiqiao and Chandra, Mohit and De Choudhury, Munmun and Kumar, Srijan and Sastry, Nishanth},
  journal={arXiv preprint arXiv:2409.19492},
  year={2024}
}

@article{asai2024openscholar,
  title={{OpenScholar}: Synthesizing Scientific Literature with Retrieval-Augmented {LMs}},
  author={Asai, Akari and He, Jacqueline and Shao, Rulin and Shi, Weijia and Singh, Amanpreet and Chang, Joseph Chee and Lo, Kyle and Soldaini, Luca and Feldman, Sergey and D'arcy, Mike and others},
  journal={arXiv preprint arXiv:2411.14199},
  year={2024}
}

@article{auer2025towards,
  title={Towards {AI}-Supported Research: A Vision of the {TIB AIssistant}},
  author={Auer, S{\"o}ren and Oelen, Allard and Jaradeh, Mohamad Yaser and Khalid, Mutahira and Keya, Farhana and Gaddipati, Sasi Kiran and D'Souza, Jennifer and Schl{\"u}ter, Lorenz and Alasti, Amirreza and Rabby, Gollam and others},
  journal={arXiv preprint arXiv:2512.16447},
  year={2025}
}

@inproceedings{azar2024general,
  title={A general theoretical paradigm to understand learning from human preferences},
  author={Azar, Mohammad Gheshlaghi and Guo, Zhaohan Daniel and Piot, Bilal and Munos, Remi and Rowland, Mark and Valko, Michal and Calandriello, Daniele},
  booktitle={International Conference on Artificial Intelligence and Statistics},
  pages={4447--4455},
  year={2024},
  organization={PMLR}
}

@inproceedings{baek2025researchagent,
  title={{ResearchAgent}: Iterative Research Idea Generation over Scientific Literature with Large Language Models},
  author={Baek, Jinheon and Jauhar, Sujay Kumar and Cucerzan, Silviu and Hwang, Sung Ju},
  booktitle={Proceedings of the 2025 Conference of the Nations of the Americas Chapter of the Association for Computational Linguistics: Human Language Technologies (Volume 1: Long Papers)},
  pages={6709--6738},
  year={2025}
}

@article{boiko2023autonomous,
  title={Autonomous chemical research with large language models},
  author={Boiko, Daniil A and MacKnight, Robert and Kline, Ben and Gomes, Gabe},
  journal={Nature},
  volume={624},
  number={7992},
  pages={570--578},
  year={2023},
  publisher={Nature Publishing Group UK London}
}

@article{chen2023theoremqa,
  title={{TheoremQA}: A Theorem-driven Question Answering Dataset},
  author={Chen, Wenhu and Yin, Ming and Ku, Max and Lu, Pan and Wan, Yixin and Ma, Xueguang and Xu, Jianyu and Wang, Xinyi and Xia, Tony},
  journal={arXiv preprint arXiv:2305.12524},
  year={2023}
}

@article{chen2024odin,
  title={{ODIN}: Disentangled Reward Mitigates Hacking in {RLHF}},
  author={Chen, Lichang and Zhu, Chen and Soselia, Davit and Chen, Jiuhai and Zhou, Tianyi and Goldstein, Tom and Huang, Heng and Shoeybi, Mohammad and Catanzaro, Bryan},
  journal={arXiv preprint arXiv:2402.07319},
  year={2024}
}

@misc{deepseek2026v4,
      title={{DeepSeek-V4}: Towards Highly Efficient Million-Token Context Intelligence}, 
      author={DeepSeek-AI},
      year={2026},
      eprint={2606.19348},
      archivePrefix={arXiv},
      primaryClass={cs.CL}
}

@article{dubois2024length,
  title={Length-Controlled {AlpacaEval}: A Simple Way to Debias Automatic Evaluators},
  author={Dubois, Yann and Galambosi, Bal{\'a}zs and Liang, Percy and Hashimoto, Tatsunori B},
  journal={arXiv preprint arXiv:2404.04475},
  year={2024}
}

@article{gaddipati2025aissistant,
  title={{AIssistant}: Human-{AI} Collaborative Review and Perspective Research Workflows in Data Science},
  author={Gaddipati, Sasi Kiran and Keya, Farhana and Rabby, Gollam and Auer, S{\"o}ren},
  journal={arXiv preprint arXiv:2509.12282},
  year={2025}
}

@article{grattafiori2024llama,
  title={The {Llama} 3 Herd of Models},
  author={Grattafiori, Aaron and Dubey, Abhimanyu and Jauhri, Abhinav and Pandey, Abhinav and Kadian, Abhishek and Al-Dahle, Ahmad and Letman, Aiesha and Mathur, Akhil and Schelten, Alan and Vaughan, Alex and others},
  journal={arXiv preprint arXiv:2407.21783},
  year={2024}
}

@article{hayes2025simulating,
  title={Simulating 500 million years of evolution with a language model},
  author={Hayes, Thomas and Rao, Roshan and Akin, Halil and Sofroniew, Nicholas J and Oktay, Deniz and Lin, Zeming and Verkuil, Robert and Tran, Vincent Q and Deaton, Jonathan and Wiggert, Marius and others},
  journal={Science},
  volume={387},
  number={6736},
  pages={850--858},
  year={2025},
  publisher={American Association for the Advancement of Science}
}

@article{hendrycks2020measuring,
  title={Measuring Massive Multitask Language Understanding},
  author={Hendrycks, Dan and Burns, Collin and Basart, Steven and Zou, Andy and Mazeika, Mantas and Song, Dawn and Steinhardt, Jacob},
  journal={arXiv preprint arXiv:2009.03300},
  year={2020}
}

@inproceedings{kim2024understanding,
  title={Understanding Users' Dissatisfaction with {ChatGPT} Responses: Types, Resolving Tactics, and the Effect of Knowledge Level},
  author={Kim, Yoonsu and Lee, Jueon and Kim, Seoyoung and Park, Jaehyuk and Kim, Juho},
  booktitle={Proceedings of the 29th International Conference on Intelligent User Interfaces},
  pages={385--404},
  year={2024}
}

@misc{laion2025ossas,
  title        = {{OSSAS} (Open Source Summaries at Scale): The {INFERENCE.NET} $\times$ {LAION} $\times$ {Grass}},
  author       = {Schuhmann, Christoph and Singh, Amarjot and Prolorenzo, Andrii and Radonjic, Andrej and Smith, Sean and Hogan, Sam},
  howpublished = {LAION Notes / Blog},
  year         = {2025},
  month        = nov,
  note         = {\url{https://laion.ai/notes/summaries/}},
}

@article{lala2023paperqa,
  title={{PaperQA}: Retrieval-Augmented Generative Agent for Scientific Research},
  author={L{\'a}la, Jakub and O'Donoghue, Odhran and Shtedritski, Aleksandar and Cox, Sam and Rodriques, Samuel G and White, Andrew D},
  journal={arXiv preprint arXiv:2312.07559},
  year={2023}
}

@article{lee2023rlaif,
  title={{RLAIF} vs. {RLHF}: Scaling Reinforcement Learning from Human Feedback with {AI} Feedback},
  author={Lee, Harrison and Phatale, Samrat and Mansoor, Hassan and Mesnard, Thomas and Ferret, Johan and Lu, Kellie and Bishop, Colton and Hall, Ethan and Carbune, Victor and Rastogi, Abhinav and others},
  journal={arXiv preprint arXiv:2309.00267},
  year={2023}
}

@article{li2024chain,
  title={Chain of Ideas: Revolutionizing Research via Novel Idea Development with {LLM} Agents},
  author={Li, Long and Xu, Weiwen and Guo, Jiayan and Zhao, Ruochen and Li, Xingxuan and Yuan, Yuqian and Zhang, Boqiang and Jiang, Yuming and Xin, Yifei and Dang, Ronghao and others},
  journal={arXiv preprint arXiv:2410.13185},
  year={2024}
}

@article{lu2024ai,
  title={The {AI} Scientist: Towards Fully Automated Open-Ended Scientific Discovery},
  author={Lu, Chris and Lu, Cong and Lange, Robert Tjarko and Foerster, Jakob and Clune, Jeff and Ha, David},
  journal={arXiv preprint arXiv:2408.06292},
  year={2024}
}

@article{m2024augmenting,
  title={Augmenting large language models with chemistry tools},
  author={M. Bran, Andres and Cox, Sam and Schilter, Oliver and Baldassari, Carlo and White, Andrew D and Schwaller, Philippe},
  journal={Nature Machine Intelligence},
  volume={6},
  number={5},
  pages={525--535},
  year={2024},
  publisher={Nature Publishing Group UK London}
}

@article{mehri2026learning,
  title={Learning User Preferences Through Interaction for Long-Term Collaboration},
  author={Mehri, Shuhaib and Kargupta, Priyanka and August, Tal and Hakkani-T{\"u}r, Dilek},
  journal={arXiv preprint arXiv:2601.02702},
  year={2026}
}

@article{merchant2023scaling,
  title={Scaling deep learning for materials discovery},
  author={Merchant, Amil and Batzner, Simon and Schoenholz, Samuel S and Aykol, Muratahan and Cheon, Gowoon and Cubuk, Ekin Dogus},
  journal={Nature},
  volume={624},
  number={7990},
  pages={80--85},
  year={2023},
  publisher={Nature Publishing Group UK London}
}

@article{omar2025multi,
  title={Multi-model assurance analysis showing large language models are highly vulnerable to adversarial hallucination attacks during clinical decision support},
  author={Omar, Mahmud and Sorin, Vera and Collins, Jeremy D and Reich, David and Freeman, Robert and Gavin, Nicholas and Charney, Alexander and Stump, Lisa and Bragazzi, Nicola Luigi and Nadkarni, Girish N and others},
  journal={Communications Medicine},
  volume={5},
  number={1},
  pages={330},
  year={2025},
  publisher={Nature Publishing Group UK London}
}

@misc{OpenAI2024GPT4omini,
  title        = {{GPT-4o} mini},
  author       = {{OpenAI}},
  year         = {2024},
  howpublished = {\url{https://openai.com}},
  note         = {OpenAI GPT-4o mini model},
}

@article{ouyang2022training,
  title={Training language models to follow instructions with human feedback},
  author={Ouyang, Long and Wu, Jeffrey and Jiang, Xu and Almeida, Diogo and Wainwright, Carroll and Mishkin, Pamela and Zhang, Chong and Agarwal, Sandhini and Slama, Katarina and Ray, Alex and others},
  journal={Advances in neural information processing systems},
  volume={35},
  pages={27730--27744},
  year={2022}
}

@article{pyzer2022accelerating,
  title={Accelerating materials discovery using artificial intelligence, high performance computing and robotics},
  author={Pyzer-Knapp, Edward O and Pitera, Jed W and Staar, Peter WJ and Takeda, Seiji and Laino, Teodoro and Sanders, Daniel P and Sexton, James and Smith, John R and Curioni, Alessandro},
  journal={npj Computational Materials},
  volume={8},
  number={1},
  pages={84},
  year={2022},
  publisher={Nature Publishing Group UK London}
}

@article{rabby2024mc,
  title={{MC-NEST}: Enhancing Mathematical Reasoning in Large Language Models Leveraging a {Monte Carlo} Self-Refine Tree},
  author={Rabby, Gollam and Keya, Farhana and Auer, S{\"o}ren},
  journal={arXiv preprint arXiv:2411.15645},
  year={2024}
}

@article{rafailov2023direct,
  title={Direct preference optimization: Your language model is secretly a reward model},
  author={Rafailov, Rafael and Sharma, Archit and Mitchell, Eric and Manning, Christopher D and Ermon, Stefano and Finn, Chelsea},
  journal={Advances in neural information processing systems},
  volume={36},
  pages={53728--53741},
  year={2023}
}

@inproceedings{rein2024gpqa,
  title={{GPQA}: A Graduate-Level {Google}-Proof {Q\&A} Benchmark},
  author={Rein, David and Hou, Betty Li and Stickland, Asa Cooper and Petty, Jackson and Pang, Richard Yuanzhe and Dirani, Julien and Michael, Julian and Bowman, Samuel R},
  booktitle={First Conference on Language Modeling},
  year={2024}
}

@article{romera2024mathematical,
  title={Mathematical discoveries from program search with large language models},
  author={Romera-Paredes, Bernardino and Barekatain, Mohammadamin and Novikov, Alexander and Balog, Matej and Kumar, M Pawan and Dupont, Emilien and Ruiz, Francisco JR and Ellenberg, Jordan S and Wang, Pengming and Fawzi, Omar and others},
  journal={Nature},
  volume={625},
  number={7995},
  pages={468--475},
  year={2024},
  publisher={Nature Publishing Group UK London}
}

@article{saito2023verbosity,
  title={Verbosity bias in preference labeling by large language models},
  author={Saito, Keita and Wachi, Akifumi and Wataoka, Koki and Akimoto, Youhei},
  journal={arXiv preprint arXiv:2310.10076},
  year={2023}
}

@article{schmidgall2025agent,
  title={Agent Laboratory: Using {LLM} Agents as Research Assistants},
  author={Schmidgall, Samuel and Su, Yusheng and Wang, Ze and Sun, Ximeng and Wu, Jialian and Yu, Xiaodong and Liu, Jiang and Liu, Zicheng and Barsoum, Emad},
  journal={arXiv preprint arXiv:2501.04227},
  year={2025}
}

@article{shao2025sciscigpt,
  title={{SciSciGPT}: Advancing Human--{AI} Collaboration in the Science of Science},
  author={Shao, Erzhuo and Wang, Yifang and Qian, Yifan and Pan, Zhenyu and Liu, Han and Wang, Dashun},
  journal={Nature Computational Science},
  pages={1--15},
  year={2025},
  publisher={Nature Publishing Group US New York}
}

@article{schulman2017proximal,
  title={Proximal policy optimization algorithms},
  author={Schulman, John and Wolski, Filip and Dhariwal, Prafulla and Radford, Alec and Klimov, Oleg},
  journal={arXiv preprint arXiv:1707.06347},
  year={2017}
}

@article{shani2024multi,
  title={Multi-turn reinforcement learning with preference human feedback},
  author={Shani, Lior and Rosenberg, Aviv and Cassel, Asaf and Lang, Oran and Calandriello, Daniele and Zipori, Avital and Noga, Hila and Keller, Orgad and Piot, Bilal and Szpektor, Idan and others},
  journal={Advances in Neural Information Processing Systems},
  volume={37},
  pages={118953--118993},
  year={2024}
}

@article{shi2024direct,
  title={Direct multi-turn preference optimization for language agents},
  author={Shi, Wentao and Yuan, Mengqi and Wu, Junkang and Wang, Qifan and Feng, Fuli},
  journal={arXiv preprint arXiv:2406.14868},
  year={2024}
}

@article{singh2025openai,
  title={{OpenAI} {GPT-5} System Card},
  author={Singh, Aaditya and Fry, Adam and Perelman, Adam and Tart, Adam and Ganesh, Adi and El-Kishky, Ahmed and McLaughlin, Aidan and Low, Aiden and Ostrow, AJ and Ananthram, Akhila and others},
  journal={arXiv preprint arXiv:2601.03267},
  year={2025}
}

@inproceedings{wang2024mint,
  title={{MINT}: Evaluating {LLMs} in Multi-turn Interaction with Tools and Language Feedback},
  author={Wang, Xingyao and Wang, Zihan and Liu, Jiateng and Chen, Yangyi and Yuan, Lifan and Peng, Hao and Ji, Heng},
  booktitle={International Conference on Learning Representations},
  volume={2024},
  pages={32593--32627},
  year={2024}
}

@article{wang2024understanding,
  title={Understanding user experience in large language model interactions},
  author={Wang, Jiayin and Ma, Weizhi and Sun, Peijie and Zhang, Min and Nie, Jian-Yun},
  journal={arXiv preprint arXiv:2401.08329},
  year={2024}
}

@article{wang2026perm,
  title={{PERM}: Psychology-grounded Empathetic Reward Modeling for Large Language Models},
  author={Wang, Chengbing and Zheng, Wuqiang and Zhang, Yang and Zhu, Fengbin and Cheng, Junyi and Xie, Yi and Wang, Wenjie and Feng, Fuli},
  journal={arXiv preprint arXiv:2601.10532},
  year={2026}
}

@article{wu2025collabllm,
  title={{CollabLLM}: From Passive Responders to Active Collaborators},
  author={Wu, Shirley and Galley, Michel and Peng, Baolin and Cheng, Hao and Li, Gavin and Dou, Yao and Cai, Weixin and Zou, James and Leskovec, Jure and Gao, Jianfeng},
  journal={arXiv preprint arXiv:2502.00640},
  year={2025}
}

@article{xiong2025toward,
  title={Toward Reliable Scientific Hypothesis Generation: Evaluating Truthfulness and Hallucination in Large Language Models},
  author={Xiong, Guangzhi and Xie, Eric and Williams, Corey and Kim, Myles and Shariatmadari, Amir Hassan and Guo, Sikun and Bekiranov, Stefan and Zhang, Aidong},
  journal={arXiv preprint arXiv:2505.14599},
  year={2025}
}

@article{yadav2021mechanism,
  title={Mechanism and Control of Meiotic {DNA} Double-Strand Break Formation in \emph{{S. cerevisiae}}},
  author={Yadav, Vikash Kumar and Claeys Bouuaert, Corentin},
  journal={Frontiers in cell and developmental biology},
  volume={9},
  pages={642737},
  year={2021},
  publisher={Frontiers Media SA}
}

@article{yamada2025ai,
  title={The {AI} Scientist-v2: Workshop-Level Automated Scientific Discovery via Agentic Tree Search},
  author={Yamada, Yutaro and Lange, Robert Tjarko and Lu, Cong and Hu, Shengran and Lu, Chris and Foerster, Jakob and Clune, Jeff and Ha, David},
  journal={arXiv preprint arXiv:2504.08066},
  year={2025}
}

@article{yang2024qwen2,
  title={{Qwen2.5} Technical Report},
  author={Yang, An and Yang, Baosong and Zhang, Beichen and Hui, Binyuan and Zheng, Bo and Yu, Bowen and Li, Chengyuan and Liu, D and Huang, F and Wei, H and others},
  journal={arXiv preprint arXiv:2412.15115},
  year={2024}
}

@inproceedings{yu2025tinyscientist,
  title={{TinyScientist}: An Interactive, Extensible, and Controllable Framework for Building Research Agents},
  author={Yu, Haofei and Xuan, Keyang and Li, Fenghai and Zhu, Kunlun and Lei, Zijie and Zhang, Jiaxun and Qi, Ziheng and Richardson, Kyle and You, Jiaxuan},
  booktitle={Proceedings of the 2025 Conference on Empirical Methods in Natural Language Processing: System Demonstrations},
  pages={558--590},
  year={2025}
}

@inproceedings{zamfirescu2023johnny,
  title={Why {Johnny} Can't Prompt: How Non-{AI} Experts Try (and Fail) to Design {LLM} Prompts},
  author={Zamfirescu-Pereira, J Diego and Wong, Richmond Y and Hartmann, Bjoern and Yang, Qian},
  booktitle={Proceedings of the 2023 CHI conference on human factors in computing systems},
  pages={1--21},
  year={2023}
}

@article{zhang2024honeycomb,
  title={{HoneyComb}: A Flexible {LLM}-Based Agent System for Materials Science},
  author={Zhang, Huan and Song, Yu and Hou, Ziyu and Miret, Santiago and Liu, Bang},
  journal={arXiv preprint arXiv:2409.00135},
  year={2024}
}

@article{zheng2023judging,
  title={Judging {LLM}-as-a-Judge with {MT-Bench} and {Chatbot Arena}},
  author={Zheng, Lianmin and Chiang, Wei-Lin and Sheng, Ying and Zhuang, Siyuan and Wu, Zhanghao and Zhuang, Yonghao and Lin, Zi and Li, Zhuohan and Li, Dacheng and Xing, Eric and others},
  journal={Advances in neural information processing systems},
  volume={36},
  pages={46595--46623},
  year={2023}
}

@article{zhou2024archer,
  title={{ArCHer}: Training Language Model Agents via Hierarchical Multi-Turn {RL}},
  author={Zhou, Yifei and Zanette, Andrea and Pan, Jiayi and Levine, Sergey and Kumar, Aviral},
  journal={arXiv preprint arXiv:2402.19446},
  year={2024}
}

\clearpage

\appendix

\begin{table*}[htbp]
\centering
\caption{Reward weight ($\gamma$) decomposition and metric definitions.}
\label{tab:reward_comparison}
\small
\resizebox{\textwidth}{!}{
\begin{tabular}{llcccp{7cm}}
\hline
\textbf{Category} & \textbf{Metric ($\phi$)} & \textbf{CollabSciReM} & \textbf{CollabLLM} & \textbf{User study} & \textbf{Description} \\
\hline

\multirow{4}{*}{$\mathcal{R}_{\text{gen}}$}
& Accuracy      & +1.3 & +1.0 & +1.0 & Correctness of technical claims and data. \\
& Completeness  & +0.9 & --   & -- & Coverage of all user sub-questions. \\
& Clarity       & +0.4 & --   & -- & Syntactic and semantic ease of understanding. \\
& Interactivity & +0.3 & +0.5 & +0.5 & Proactive engagement and user feedback. \\
\hline

\multirow{3}{*}{$\mathcal{R}_{\text{sci}}$}
& Justification & +0.8 & -- & -- & Provision of evidence, citations, or derivations. \\
& Logical Flow  & +0.6 & -- & -- & Internal consistency and step-by-step reasoning. \\
& Novelty       & +0.3 & -- & -- & Identification of unique scientific insights. \\
\hline

\multirow{2}{*}{$\mathcal{R}_{\text{struct}}$}
& Format      & +0.2 & --   & -- & Adherence to \LaTeX{}, Markdown, or schemas. \\
& Token Count & -0.4 & -0.5 & -- & Penalty per token unit ($\lambda = 0.001$). \\
\hline

\end{tabular}}
\end{table*}

\begin{table*}[htbp]
\centering
\caption{Hyperparameters for SFT and IPO training stages.}
\label{tab:hyperparams}
\small
\resizebox{\textwidth}{!}{
\begin{tabular}{@{}lll@{}}
\toprule

\textbf{Hyperparameter} & \textbf{SFT} & \textbf{IPO} \\
\midrule

\textit{Hardware \& Infrastructure} & & \\
GPUs                & 4 $\times$ NVIDIA H100 (80GB) & 4 $\times$ NVIDIA H100 (80GB) \\
Precision           & bfloat16                      & bfloat16 \\
Parallelism         & DeepSpeed ZeRO-2              & DeepSpeed ZeRO-2 \\
\midrule

\textit{LoRA Configuration} & & \\
LoRA Rank ($r$)       & 32                & 32 \\
LoRA Alpha ($\alpha$) & 16                & 16 \\
Target Modules        & All Linear Layers & All Linear Layers \\
Quantization          & 4-bit (NF4)       & None (bfloat16) \\
\midrule

\textit{Training Dynamics} & & \\
Learning Rate       & $1 e{-5}$ & $5 e{-6}$ \\
LR Scheduler        & Cosine              & Cosine \\
Max Gradient Norm   & 1.0                 & 1.0 \\
Training Epochs     & 7                   & 8 \\
\midrule

\textit{Batch Configuration} & & \\
Per-device Batch Size & 4  & 4 \\
Gradient Accumulation & 4  & 4 \\
Effective Batch Size  & 16 & 16 \\
\midrule

\textit{Objective \& Length} & & \\
Loss Type           & Cross-Entropy     & IPO ($\beta=0.2$) \\
Max Sequence Length & 4096              & 4096 \\
Max Prompt Length   & --                & 4096 \\

\bottomrule
\end{tabular}}
\end{table*}

\section{Reward Model Configurations and Reproducibility}
\label{appendix: sci_metrics_and_weights}

To ensure the reproducibility of \textsc{CollabSciReM}, we detail the weighting coefficients $\gamma$ used to decompose the reward function $\mathcal{R}^*(\tau \mid g)$. Each metric score $\phi_i$ is generated by a critic model (\textit{gpt-5-mini}). The scores are discretized to the set $\{0, 0.5, 1\}$ for all metrics, except accuracy, which is restricted to binary values $\{0,1\}$.

We began by considering a broad set of scientific objectives alongside general conversational rewards, due to the lack of prior works on science-aware reward models. Through iterative experimentation and empirical validation, we refined this set by removing redundant or conflicting components, resulting in the final decomposition and weighting presented in Table~\ref{tab:reward_comparison}. The total reward is calculated as a weighted sum across three primary categories: General Conversational Utility ($\mathcal{R}_{\text{gen}}$), Scientific Rigor ($\mathcal{R}_{\text{sci}}$), and Structural Efficiency ($\mathcal{R}_{\text{struct}}$).

\section{Experimental Setup and Hyperparameters}
\label{appendix:hyperparams}

\begin{figure*}[t]
    \centering
    \includegraphics[width=0.95\textwidth]{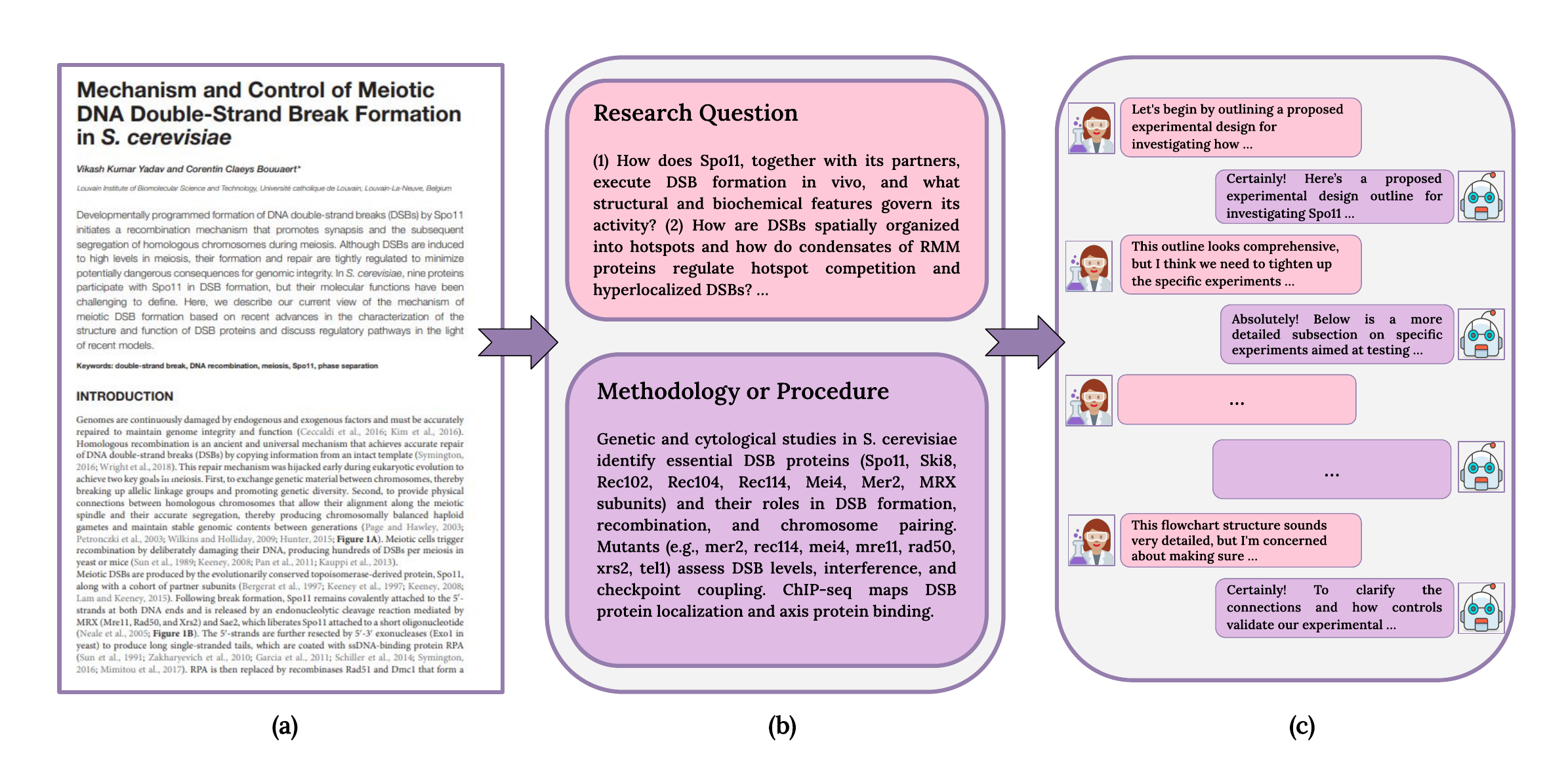}
    \caption{Synthetic dataset generation pipeline from a research paper to a conversational excerpt. (a) A research paper (excerpt from \citep{yadav2021mechanism}) is transformed into (b) a structured representation in which research questions and methodologies are extracted. (c) These components are formalized into conversational excerpts, with user queries grounded in research questions and model responses derived from the associated methodologies.}
    \label{fig:synthetic_data_excerpt}
\end{figure*}

Figure~\ref{fig:synthetic_data_excerpt} showcases our pipeline from converting research papers to conversational excerpts. Furthermore,
we conduct our experiments on the \textsc{Llama} and \textsc{Qwen} model families. Training is split into two phases: Supervised Fine-Tuning (SFT) followed by offline preference optimization with the Identity Preference Optimization (IPO) loss \citep{azar2024general}, which is less prone than DPO to overfitting deterministic preferences. Table~\ref{tab:hyperparams} showcases the relevant hyperparameters used in the experiment phase.

\section{User Study Setup}
\label{appendix: user_study_setup}

\begin{figure*}[t]
    \centering

    \begin{subfigure}[b]{0.32\textwidth}
        \centering
        \includegraphics[width=\linewidth,height=4.5cm,keepaspectratio]{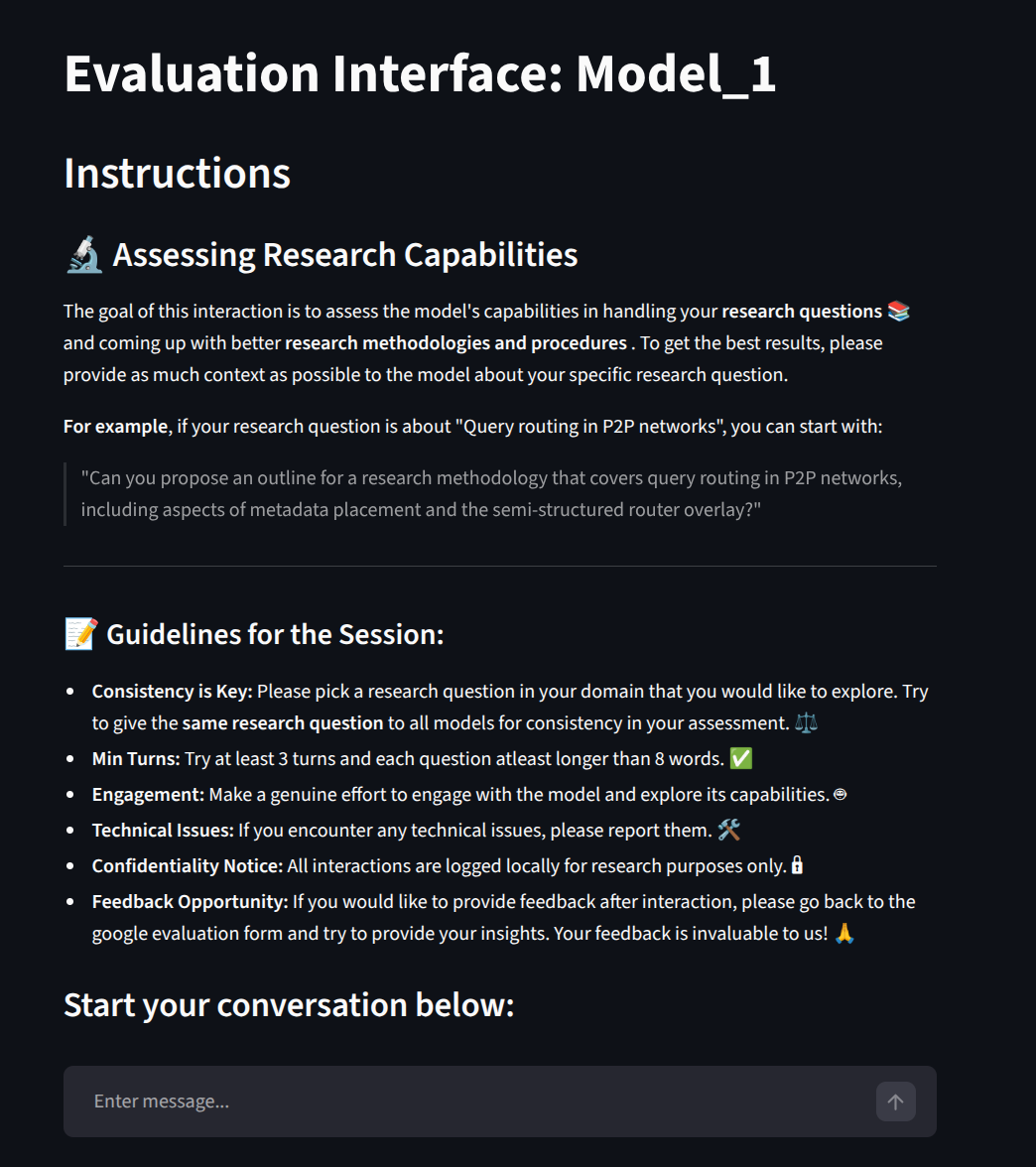}
        \caption{Model Interface}
        \label{fig:user_study_model_interface}
    \end{subfigure}
    \hfill
    \begin{subfigure}[b]{0.30\textwidth}
        \centering
        \includegraphics[width=\linewidth,height=4.5cm,keepaspectratio]{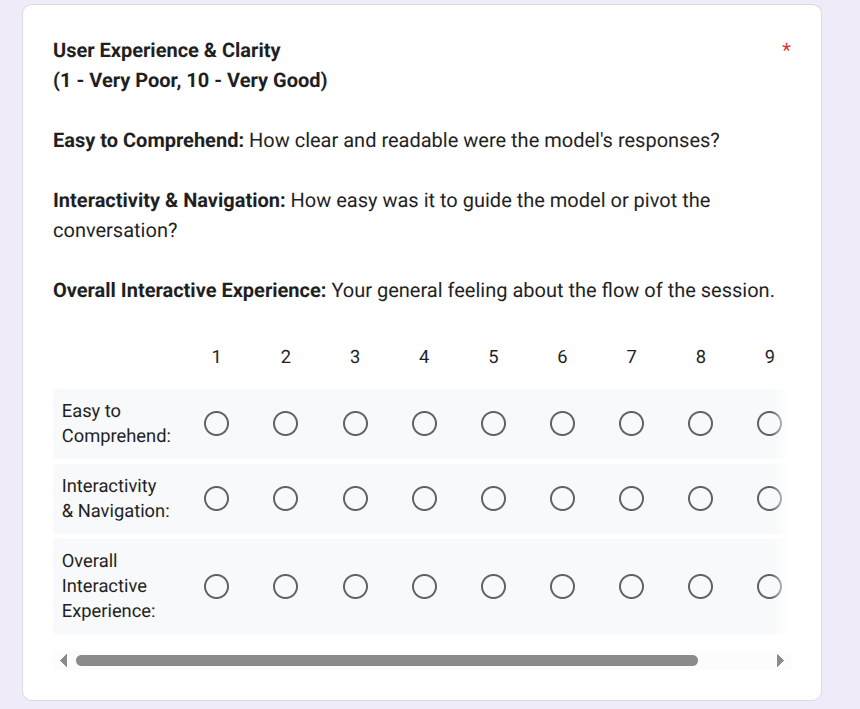}
        \caption{Interactive Experience Evaluation}
        \label{fig:user_study_interactive_experience}
    \end{subfigure}
    \hfill
    \begin{subfigure}[b]{0.30\textwidth}
        \centering
        \includegraphics[width=\linewidth,height=4.5cm,keepaspectratio]{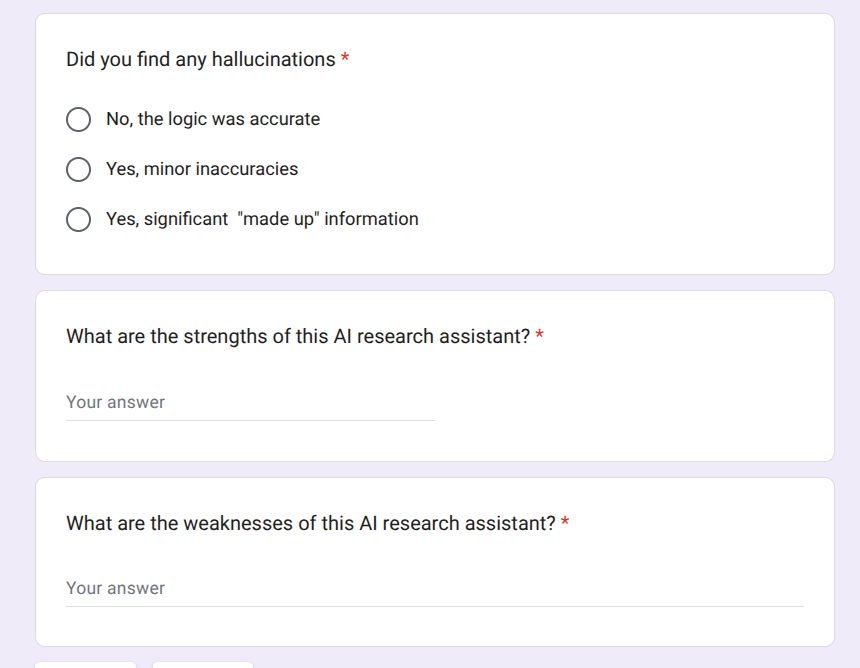}
        \caption{Hallucinations Evaluation}
        \label{fig:user_study_hallucinations}
    \end{subfigure}

    \caption{Snippets from the user study evaluation interface.}
    \label{fig:user_study_4subfigures}
\end{figure*}

We conducted a user study with evaluators drawn from diverse domain experts. All participants ($N=25$) had at least a graduate degree and sufficient experience in evaluating AI-related tasks. Figure \ref{fig:user_study_4subfigures} presents the screenshots of the model's interactive interface and evaluation form, and Table \ref{tab:qualitative_feedback} showcases the respective strengths and weaknesses of the base \textsc{Llama-3.1-8B-Instruct}, general-reward variant and \textsc{CollabSciReM}, as informed by the human evaluators.

Qualitative feedback is summarized in Table~\ref{tab:qualitative_feedback}. Participants found the base model well structured but weak in scientific reasoning and unable to propose new research directions. The general-reward variant improved fluency and formatting but required extensive prompting to produce novel insights or rigorous methodology. \textsc{CollabSciReM} produced structured, methodology-focused responses with novel and specialized research hypotheses, although some participants found its responses occasionally too concise.

\begin{table*}[t]
\centering
\caption{Qualitative Feedback from Human Evaluators.}
\label{tab:qualitative_feedback}
\resizebox{\textwidth}{!}{%
\begin{tabular}{l|l|l}
\hline
\rowcolor[HTML]{EFEFEF}
\multicolumn{1}{c|}{\cellcolor[HTML]{EFEFEF}{\color[HTML]{1F1F1F} \textbf{Model}}} & \multicolumn{1}{c|}{\cellcolor[HTML]{EFEFEF}{\color[HTML]{1F1F1F} \textbf{Strengths}}}                                                                                                                                                                       & \multicolumn{1}{c}{\cellcolor[HTML]{EFEFEF}{\color[HTML]{1F1F1F} \textbf{Weaknesses}}}                                                                                                                                                                                                                                            \\ \hline
{\color[HTML]{1F1F1F} Llama-3.1-8B-Instruct}                                                         & {\color[HTML]{1F1F1F} \textit{\begin{tabular}[c]{@{}l@{}}``Clear and answered the questions appropriately.",\\ ``Strong structure and clarity."\end{tabular}}}                                                                                             & {\color[HTML]{1F1F1F} \textit{\begin{tabular}[c]{@{}l@{}}``Flawed reasoning.", ``The model's approach was generic \\ and lacked innovation, ...", ``The model provided too many \\ options to try, and its initial response was not thorough at all. \\ It felt as though I had to pry additional steps out of it."\end{tabular}}} \\ \hline
{\color[HTML]{1F1F1F} General-reward Variant}                                            & {\color[HTML]{1F1F1F} \textit{\begin{tabular}[c]{@{}l@{}}``Decent response formatting." , ``Clear, concise",\\  ``Clear language flow and formatting."\end{tabular}}}                                                                                       & {\color[HTML]{1F1F1F} \textit{\begin{tabular}[c]{@{}l@{}}``I had to point it more specifically at research methodology\\  over answering the question.", ``It got some technical \\ information wrong..", ``Limited novelty."\end{tabular}}}                                                                                        \\ \hline
{\color[HTML]{1F1F1F} CollabSciReM}                                                 & {\color[HTML]{1F1F1F} \textit{\begin{tabular}[c]{@{}l@{}}``It was easy to understand; well structured and I \\liked  how it was straight to the point and just  gave \\ an overview.", ``Clear, structured, and readable"\end{tabular}}} & {\color[HTML]{1F1F1F} \textit{\begin{tabular}[c]{@{}l@{}}``While methodologically solid, the responses rely heavily  on \\  standard qualitative techniques.", ``Too concise..."\end{tabular}}}    \\ \hline
\end{tabular}%
}
\end{table*}

\begin{table*}[t]
\centering
\caption{Performance of 3B models on (a) multi-turn benchmarks (MT-Bench-STEM, MINT) and (b) single-turn benchmarks (GPQA-D, MMLU-Valid, TheoremQA). Metrics and evaluation settings follow Table~\ref{tab:evaluation_results}. Results are mean $\pm$ sample standard deviation over three seeds; bold values indicate the best performance in each column within each model family.}
\label{tab:evaluation_results_3b}
\small
\setlength{\tabcolsep}{3.5pt}
\resizebox{\textwidth}{!}{%
\begin{tabular}{@{}l *{9}{c}@{}}
\toprule
\multicolumn{10}{l}{\textit{(a) Multi-turn benchmarks}} \\
\midrule
& \multicolumn{6}{c}{\textbf{MT-Bench-STEM}} & \multicolumn{3}{c}{\textbf{MINT}} \\
\cmidrule(lr){2-7} \cmidrule(lr){8-10}
\textbf{Model} & \textbf{Overall} ($\uparrow$) & \textbf{Acc.} ($\uparrow$) & \textbf{Format} ($\uparrow$) & \textbf{Useful.} ($\uparrow$) & \textbf{Relev.} ($\uparrow$) & \textbf{TC} ($\downarrow$) & \textbf{SR@5 (\%)} ($\uparrow$) & \textbf{TC} ($\downarrow$) & \textbf{APT} ($\uparrow$) \\
\midrule
Llama-3.2-3B-Instruct &
\result{6.40}{0.38} & \result{6.02}{0.46} & \result{7.82}{0.20} & \textbf{\result{6.25}{0.48}} & \result{8.12}{0.29} & \result{447}{26} &
\textbf{\result{35.56}{1.94}} & \result{265.06}{20.44} & \result{135.01}{17.10} \\
Llama-3.2-3B-CollabLLMSci &
\textbf{\result{6.48}{0.38}} & \textbf{\result{6.06}{0.42}} & \textbf{\result{7.93}{0.24}} & \result{6.15}{0.25} & \result{8.24}{0.15} & \textbf{\result{344}{9}} &
\result{31.85}{2.89} & \textbf{\result{174.44}{4.47}} & \result{182.94}{21.00} \\
Llama-3.2-3B-CollabSciReM &
\result{6.24}{0.20} & \result{5.83}{0.24} & \result{7.82}{0.09} & \result{5.98}{0.21} & \textbf{\result{8.26}{0.13}} & \result{358}{16} &
\result{35.41}{2.60} & \result{181.63}{3.06} & \textbf{\result{195.13}{17.35}} \\
\midrule
Qwen-2.5-3B-Instruct &
\textbf{\result{7.44}{0.06}} & \textbf{\result{7.15}{0.02}} & \textbf{\result{8.54}{0.05}} & \textbf{\result{7.22}{0.09}} & \textbf{\result{8.82}{0.06}} & \result{521}{23} &
\result{52.15}{0.93} & \result{287.66}{1.29} & \result{181.29}{3.24} \\
Qwen-2.5-3B-CollabLLMSci &
\result{6.95}{0.48} & \result{6.63}{0.55} & \result{8.24}{0.13} & \result{6.56}{0.43} & \result{8.36}{0.09} & \textbf{\result{274}{4}} &
\result{49.93}{3.37} & \textbf{\result{153.18}{1.09}} & \textbf{\result{325.84}{19.58}} \\
Qwen-2.5-3B-CollabSciReM &
\result{6.85}{0.15} & \result{6.67}{0.22} & \result{8.06}{0.16} & \result{6.61}{0.01} & \result{8.18}{0.04} & \result{275}{3} &
\textbf{\result{54.37}{1.36}} & \result{172.64}{3.69} & \result{315.11}{13.31} \\
\midrule
\multicolumn{10}{l}{\textit{(b) Single-turn benchmarks}} \\
\midrule
& \multicolumn{3}{c}{\textbf{GPQA-D}} & \multicolumn{3}{c}{\textbf{MMLU-Valid}} & \multicolumn{3}{c}{\textbf{TheoremQA}} \\
\cmidrule(lr){2-4} \cmidrule(lr){5-7} \cmidrule(lr){8-10}
\textbf{Model} & \textbf{Acc (\%)} ($\uparrow$) & \textbf{TC} ($\downarrow$) & \textbf{APT} ($\uparrow$) & \textbf{Acc (\%)} ($\uparrow$) & \textbf{TC} ($\downarrow$) & \textbf{APT} ($\uparrow$) & \textbf{Acc (\%)} ($\uparrow$) & \textbf{TC} ($\downarrow$) & \textbf{APT} ($\uparrow$) \\
\midrule
Llama-3.2-3B-Instruct &
\result{16.84}{1.86} & \result{1520.46}{59.89} & \result{11.07}{1.14} &
\result{24.82}{1.24} & \result{640.31}{10.50} & \result{38.75}{1.51} &
\textbf{\result{14.28}{0.49}} & \result{965.50}{51.00} & \result{14.83}{0.97} \\
Llama-3.2-3B-CollabLLMSci &
\textbf{\result{25.93}{0.58}} & \textbf{\result{738.49}{10.21}} & \textbf{\result{35.11}{0.52}} &
\textbf{\result{32.48}{0.69}} & \textbf{\result{409.53}{0.67}} & \textbf{\result{79.32}{1.57}} &
\result{13.39}{0.61} & \textbf{\result{447.21}{12.80}} & \textbf{\result{29.92}{0.55}} \\
Llama-3.2-3B-CollabSciReM &
\result{23.23}{1.49} & \result{780.55}{26.91} & \result{29.76}{1.63} &
\result{30.26}{0.17} & \result{416.18}{2.08} & \result{72.72}{0.77} &
\result{13.39}{0.58} & \result{447.77}{7.96} & \result{29.89}{1.11} \\
\midrule
Qwen-2.5-3B-Instruct &
\result{24.75}{2.30} & \result{1006.88}{26.90} & \result{24.65}{2.91} &
\result{50.64}{0.51} & \result{580.51}{0.36} & \result{87.23}{0.84} &
\textbf{\result{16.47}{0.58}} & \result{859.60}{35.30} & \result{19.22}{1.49} \\
Qwen-2.5-3B-CollabLLMSci &
\result{29.46}{2.39} & \result{412.61}{3.46} & \result{71.44}{6.33} &
\result{50.88}{1.21} & \result{310.65}{0.29} & \result{163.79}{3.96} &
\result{15.08}{0.78} & \result{367.02}{4.74} & \textbf{\result{41.08}{1.68}} \\
Qwen-2.5-3B-CollabSciReM &
\textbf{\result{30.47}{0.86}} & \textbf{\result{394.78}{2.77}} & \textbf{\result{77.19}{2.23}} &
\textbf{\result{52.38}{0.51}} & \textbf{\result{295.49}{1.43}} & \textbf{\result{177.28}{1.97}} &
\result{13.52}{0.38} & \textbf{\result{339.77}{7.24}} & \result{39.79}{0.39} \\
\bottomrule
\end{tabular}%
}
\end{table*}
\section{Evaluation on 3B models}
\label{appendix: 3b_results}

Table~\ref{tab:evaluation_results_3b} reports results for \textsc{Llama-3.2-3B-Instruct} and \textsc{Qwen-2.5-3B-Instruct} under the same training and evaluation setup as the 7B and 8B models. At this scale, the effect of fine-tuning on efficiency carries over: \collabscirem uses 20--61\% fewer tokens than the base models across all ten settings and improves accuracy on GPQA-D (\textsc{Llama}: 16.84\% $\rightarrow$ 23.23\%; \textsc{Qwen}: 24.75\% $\rightarrow$ 30.47\%) and MMLU-Validation (\textsc{Llama}: 24.82\% $\rightarrow$ 30.26\%; \textsc{Qwen}: 50.64\% $\rightarrow$ 52.38\%). As with the larger models, both fine-tuned variants fall below the base models on TheoremQA. The advantage of the science-aware reward over \collabllmsci, however, is less consistent at 3B. With \textsc{Qwen}, \collabscirem leads on GPQA-D, MMLU-Validation, and MINT, where it is also the only variant to exceed the base model (54.37\% vs.\ 52.15\% SR@5). With \textsc{Llama}, it leads only on MINT, ties on TheoremQA, and trails \collabllmsci on GPQA-D, MMLU-Validation, and MT-Bench-STEM. On MT-Bench-STEM, neither fine-tuned \textsc{Qwen} model matches the base model. We hypothesize that smaller models have less capacity to benefit from the additional fine-grained reward terms, which mainly reward elaborated reasoning, while the structural reward's compression effect is easier to learn. The near-chance MMLU-Validation accuracy of the \textsc{Llama} base model (24.82\%) also suggests that, at this scale, part of the accuracy gain from fine-tuning reflects more reliable answer formatting rather than improved reasoning. Overall, the 3B results support the efficiency benefits of science-aware fine-tuning but indicate that its accuracy advantage over a general collaborative reward depends on model scale and family.

\newpage
\section{Scientific-Quality Evaluation: Judge Prompt, Rubric and Example}
\label{app:sciq-judge}

\paragraph{Rubric.} Table~\ref{tab:sciq-rubric} defines the dimensions.
\begin{table}[ht]
\centering
\small
\begin{tabular}{@{}>{\raggedright\arraybackslash}p{0.29\linewidth}p{0.65\linewidth}@{}}
\toprule
\textbf{Dimension} & \textbf{Definition} \\
\midrule
Methodological soundness & Assesses whether the proposed study design is valid, with appropriate controls, careful handling of confounds, sound measurement and analysis, and a feasible plan. \\
Justification & Assesses whether each design choice is supported by specific, checkable reasons. Citations count only if they identify a real, relevant work, and invented, vague or unverifiable references count against the response. \\
Logical consistency & Assesses whether the response is free of contradictions within and across turns and whether later turns build coherently on earlier ones. \\
Factual correctness & Assesses the technical and factual accuracy of the response. The judge lists the concrete errors it finds for each side. \\
\midrule
Novelty & Assesses whether the response offers non-obvious, useful ideas that go beyond a textbook answer. \\
Responsiveness & Assesses how well the assistant addresses the researcher's evolving requests and questions across turns. \\
Overall & Asks which assistant a domain expert would rather collaborate with.  \\
\bottomrule
\end{tabular}
\caption{Scientific-quality rubric.}
\label{tab:sciq-rubric}
\end{table}

\paragraph{Scoring.} Each pair is judged twice, with the assistants' positions swapped. Per dimension, a model wins a prompt (score 1) only if it is preferred in both orders, loses (score 0) if the other model is preferred in both, and ties (score 0.5) otherwise, including when the two verdicts disagree. The win rate is the mean score over the 300 held-out prompts.

\paragraph{Example.} A comparison of CollabSciReM and \collabllmsci (both Qwen-2.5-7B) on a held-out medical-imaging paper (\emph{Evaluation of Total Variation Noise-Reduction Algorithm for PET/MR Images Using MR-AC Dixon-CAIPI and MR-AC Dixon-GRAPPA}). Both conversations open with the same researcher message asking for help designing a study comparing a total-variation denoising algorithm with Wiener and median filters on PET image quality (CNR, COV). Excerpts:

\begin{quote}\small
\textbf{Researcher (turn 3, CollabLLM-science conversation):} ``Can you identify key studies that have reported optimal parameters for both Dixon-CAIPI and Dixon-GRAPPA in PET/MR imaging?''\\
\textbf{CollabLLM-science:} ``Review studies like `Optimizing Image Quality in MR-AC Using CAIPI' published in Radiology, which suggests TR/TE settings of 2800ms/16.2ms [\dots]''\\
\textbf{CollabLLM-science (turn 4):} ``\dots Radiology, Vol.~295, No.~2, 2020. DOI: [Insert DOI here]''

\medskip
\textbf{Researcher (turn 3, CollabSciReM conversation):} ``How can we determine the expected effect sizes and variability for our measures (CNR and COV) [\dots] how can we assess inter-rater reliability among the radiologists?''\\
\textbf{CollabSciReM:} ``Review similar PET studies [\dots] Note the standard deviations and mean differences. [\dots] If available, use pilot data to estimate variability. [\dots]'' (followed by a power analysis, a mixed-effects model specification and an ICC-based reliability assessment)
\end{quote}

\paragraph{Judge.} DeepSeek-V4-Pro (\texttt{deepseek/deepseek-v4-pro-0813}, via OpenRouter, temperature 0) compares two assistants' four-turn conversations about the same held-out research question. The placeholders \texttt{\{question\}}, \texttt{\{reference\}}, \texttt{\{conv\_a\}} and \texttt{\{conv\_b\}} are filled with the paper's research question, the methodology reported in the original paper, and the two conversations, with model identities removed.

\begin{tcolorbox}[promptbox, title={Judge Prompt}]
\begin{Verbatim}
You are an expert scientific reviewer. Two AI research assistants (A and B) each held a four-turn conversation with a researcher who is planning a study. The researcher's later messages differ slightly because they react to each assistant; judge only the assistants' contributions.

Researcher's research question:
<<<
{question}
>>>

Reference methodology used by the original study (one valid approach, not the only acceptable one: do not penalize sound alternatives, but use it to check domain correctness and feasibility):
<<<
{reference}
>>>

Conversation with assistant A:
<<<
{conv_a}
>>>

Conversation with assistant B:
<<<
{conv_b}
>>>

Compare A and B on each dimension and pick "A", "B" or "tie" (tie only if they are genuinely equivalent). Judge substance, not length or formatting: more text is better only if it adds valid scientific content.

1. methodological_soundness: validity of the proposed study design - controls, confounds, measurement, analysis, feasibility.
2. justification: specific, checkable reasons for each design choice. A citation earns credit only if it identifies a real, relevant work you are confident exists; invented, vague or unverifiable references count against the response.
3. logical_consistency: no contradictions within or across turns; later turns build coherently on earlier ones.
4. factual_correctness: technical and factual accuracy; list concrete errors for each side.
5. novelty: non-obvious, useful ideas beyond a textbook answer.
6. responsiveness: addresses the researcher's evolving requests and questions.
7. overall: which assistant a domain expert would rather collaborate with.

Respond with a single JSON object and nothing else, in this form:
{"methodological_soundness": {"winner": "A", "reason": "one sentence"},
 "justification": {"winner": "B", "reason": "one sentence"},
 "logical_consistency": {"winner": "tie", "reason": "one sentence"},
 "factual_correctness": {"winner": "A", "reason": "one sentence"},
 "novelty": {"winner": "A", "reason": "one sentence"},
 "responsiveness": {"winner": "B", "reason": "one sentence"},
 "overall": {"winner": "A", "reason": "one sentence"},
 "factual_errors": {"A": ["..."], "B": ["..."]}}
\end{Verbatim}

\end{tcolorbox}

\section{Prompts}
\label{appendix: prompts}

In this section, we present the prompts used for synthetic data generation and reward metric computation, adapted from \citet{wu2025collabllm} and further refined for the scientific contexts.

\subsection{User Simulation Prompt}

\begin{tcolorbox}[promptbox, title={User Simulator Prompt}]
\begin{Verbatim}
You are role-playing as a human USER interacting with an AI collaborator to complete a specific task. Your goal is to generate realistic, natural responses that a user might give in this scenario.

## Input Information:
You will be provided with:
- Task Description: The type of task you are trying to accomplish.
- Complete Prompt or Reference Goal: This field may include the complete user request/query or a reference answer to user's request. Use this field to understand the user's intent, requirements, or what would count as a satisfactory outcome.
- Chat History: The ongoing conversation between you (as the user) and the AI

Inputs:
<|The Start of Task Description (Not visible to the AI)|>
{task_desc}
<|The End of Task Description|>

<|The Start of Complete Prompt or Reference Goal (Not visible to the AI)|>
{single_turn_prompt}
<|The End of Complete Prompt or Reference Goal|>

<|The Start of Chat History|>
{chat_history}
<|The End of Chat History|>

## Guidelines:
- Stay in Character: Role-play as a human USER. You are NOT an AI. Maintain a consistent personality throughout the chat.
- Minimize Effort: IMPORTANT! As a user, avoid being too detailed in your responses. Provide vague or incomplete demands in the early stages of the conversation to minimize your effort. Let the AI ask for clarification rather than providing everything upfront.
- Knowledge Background: Reflect the user's knowledge level in the role-playing. If the user is less knowledgeable about a task, they might not notice incorrect statements. Ask questions that demonstrate your current understanding and areas of confusion.
- Occasionally Make Mistakes: Real-world users might misspell words, provide incorrect dates, give wrong information, or ask unclear questions. Simulate this behavior to reflect natural interactions.
- Mention Personal Preferences: Include preferences or constraints that might influence your requests or responses. For example, "I prefer short answers," "I need this done quickly," or "I like detailed comments in code."
- Goal-Oriented: Keep the chat focused on your intent. Avoid small talk or digressions. Redirect the chat back to the main objective if it starts to stray.

## Output Format:
You should output a JSON object with three entries:
- "current_answer" (str): Briefly summarize the AI's current solution to the task.
- "thought" (str): Output your thought process as a user deciding what to say next. Consider:
  1. Have you obtained a satisfactory solution from the AI? If yes, you can terminate this chat.
  2. If not, what specific part of the problem or solution are you struggling with?
  3. Has the AI asked you to perform a task or answer a question? If so, how should you approach it?
  4. Are you noticing any patterns or potential misunderstandings that need clarification?
  5. If you're stuck, how can you phrase your question to get the most helpful response while demonstrating your current understanding?
- "response" (str): Based on your thought process, respond to the AI as the user you are role-playing. Stop immediately when the user's response is completed.

## Important Notes:
- Respond Based on Previous Messages: Your responses should be based on the context of the current chat history. Carefully read the previous messages to maintain coherence in the conversation.
- Conversation Flow: If "Current Chat History" is empty, start the conversation from scratch with an initial request. Otherwise, continue based on the existing conversation.
- Don't Copy Input Directly: Use the provided information for understanding context only. Avoid copying target queries or any provided information directly in your responses.
- Completion Signal: Use "{terminal_signal}" as your response when you believe your goal has been solved or if you determine the AI cannot help further.
- Double check if the JSON object is formatted correctly. Ensure that all fields are present and properly structured.

Remember to stay in character as a user throughout your response, and follow the instructions and guidelines carefully. 
\end{Verbatim}

\end{tcolorbox}

\subsection{Teacher Model Prompt}
\begin{tcolorbox}[promptbox, title={Teacher Model}]
\begin{Verbatim}
You are an AI assistant interacting with a user to perform tasks such as methodology generation, implementation planning and expected scientific result outcomes. Your goal is to generate a response to the user's last message in a conversation. You should be helpful, collaborative, and highly interactive.

I will provide you with the following information:
- Conversation History: This is the complete chat history where you need to respond to the last user message.
- Additional Information (Optional): This may include reference knowledge with a question and answer to give you relevant context.

<|The Start of Conversation History|>  
{chat_history}
<|The End of Conversation History|>

<|The Start of Additional Information|>
{additional_info}
<|The End of Additional Information|>

# Guidelines:
1. Understanding and Engagement
   - Accurately interpret the user's research problem throughout the conversation.
   - Acknowledge previous interactions to maintain scientific context and continuity in the conversation.

2. Interactivity (Important!)
   - Ask clarifying questions if the user's request lacks detail or is ambiguous. Such as the limitations of theirs, specific function format for a coding task, or the context of the domain.
   - Ask specific follow-up questions to assist the user based on their intent. Avoid general questions like "Do you have any further questions? Let me know." Instead, focus on specifics like, "Would you like to focus more on X?" or "Can you clarify your limitations for Y?"
   - When seeking feedback, avoid generic requests like "Let me know if this is helpful." Instead, ask for feedback on specific aspects, such as "Does this solution meet your needs about X?"
   - Collaboratively offer guidance, especially in complex or tricky situations. Provide specific suggestions on potential next steps.
   - Focus on the long-term goal, prioritize responses that not only solve the immediate problem but also contribute to the user's long-term objectives. Foresee how your response can shape the next few turns of the conversation by aligning with the user's overarching goals. 

3. Efficiency and User Consideration
   - Be mindful of how much the user needs to read or type, keeping the interaction concise and focused.
   - When asking for feedback or presenting options, provide multiple-choice suggestions or specific prompts to make it easier for the user to respond quickly.
   - Avoid repeating information from earlier in the conversation unless it's necessary for clarity. Ensure your responses are not redundant.

4. Communication Style
   - Be honest in your responses. If you are unsure of something, say, "I cannot come up with a solution" and suggest ways the user could find the information.
   - Ensure your responses are clear, well-structured, referenced and free from grammatical errors.

# Output Format:
You should output a JSON object with three entries:
- "current_problem" (str): What is the current problem the user is asking and what are the solutions they might be interested in?
- "thought" (str): Output your thought process deciding what to say next. You may consider the following: 
   1. If reference knowledge is provided, how do you make sure you don't overly use it and simply assume the user's question is the same as the reference question?
   2. What information is missing from the user's input? Does the user's message lack any necessary details?
   3. Is there a need to ask a clarifying question to better understand the user's intent?
   4. Does the user seem confused or unclear on a particular topic? How can you address that confusion?
   5. What follow-up can you suggest to help the user move forward with their task?
   6. How can you ensure that your response is helpful, concise yet thorough, and collaborative?
   7. Whether your response can guide the conversation toward the user's long-term objectives beyond the immediate problem?
- "response" (str): Based on your thought process and chat history, provide your response following the guidelines to the user. Keep your response within {max_new_tokens} tokens to avoid being cut off. 

# Notes:
- Clarifying Questions: If the user's message is unclear or lacks necessary details, always ask for clarification rather than making assumptions. Ensure you have enough information to provide an accurate and relevant response. For example, if the user asks, "Can you solve this equation?" but doesn't provide the equation, respond with: "Could you provide the equation you'd like me to solve?"
- Reference Knowledge Usage: If reference knowledge is provided in the additional information, use it as context but do not assume that the user's question will exactly match the reference question. Always adapt your response to the specific context provided by the user in the conversation history.
- Ensuring Interactivity: Encourage more interaction with the user by engaging in at least three conversational turns. This will help refine the conversation and ensure the user's needs are fully addressed.
- Double check if the JSON object is formatted correctly. Ensure that all fields are present and properly structured.

Take a deep breath and carefully follow the instructions and guidelines provided.
\end{Verbatim}
\end{tcolorbox}

\subsection{Critic/Judge LLM Scoring Prompts}

\begin{tcolorbox}[promptbox, title={Accuracy Scoring}]
\begin{Verbatim}
You are a helpful and meticulous evaluator. Your task is to evaluate the *accuracy* of an AI model's answer to a target question.

You will be given the target question, the ground truth answer, and the model's response.

Provided Information:

<|The Start of Target Question and Ground Truth Answer|> Target Question: {single_turn_prompt} Ground Truth Answer: {groundtruth}
<|The End of Target Question and Ground Truth Answer|>

<|The Start of The Model's Response|>
{completion}
<|The End of The Model's Response|>

You should determine whether the model's final response to the target question is factually correct and consistent with the provided ground truth.

Rating criteria (binary):
  • 1 = Correct   — the response matches the ground truth. • 0 = Incorrect — the response contradicts or misses the ground truth.

Output format (JSON):
{{
    "thought": "<your reasoning here>",
    "accuracy": <0 or 1>
}}

Double check if the JSON object is formatted correctly. Ensure that all fields are present and properly structured. Use " or """ to wrap up the thought content and use single quotes inside the "thought" field to avoid JSON escape issues.

Your evaluation:
\end{Verbatim}
\end{tcolorbox}

\begin{tcolorbox}[promptbox, title={Completeness Scoring}]
\begin{Verbatim}
You are a helpful and meticulous research methodology evaluator. 
Your task is to evaluate the *completeness* of the responses provided by an AI 
assistant to the user questions in a given conversation:

<|The Start of the Conversation to be Evaluated|>
{chat_history}
<|The End of the Conversation to be Evaluated|>

You should assess the assistant's response on completeness to the user's 
needs.

Give a float number between 0 and 1, where:
    1 = Totally complete: The assistant's responses are comprehensive and 
    fully address all aspects of the user's question.
     - Example: The assistant thoroughly understands the user's question and 
     provides a detailed, accurate response that directly addresses all parts 
     of the user's intent, such as "Based on your question about climate 
     change, here are the key factors contributing to it..."
    0.5 = Moderately complete: The assistant's responses are somewhat complete but may miss some key aspects or provide partially relevant information.
     - Example: The assistant provides a response that addresses part of the user's question but misses some important details, such as "Climate change is influenced by various factors," without elaborating on the specifics.
    0 = Low completeness: The assistant's responses show little to no completeness with respect to the user's needs, providing incomplete or inaccurate information that does not fully address the user's questions.
     - Example: The assistant provides a vague or off-topic response that does not relate to the user's question, such as "Climate change is bad," without any further context or relevance to the user's inquiry.

Output format (JSON):
{{
    "thought": "<How complete is the assistant?>",
    "completeness": <score>
}}

Double check if the JSON object is formatted correctly. Ensure that all fields are present and properly structured. Use " or """ to wrap up the thought content and use single quotes inside the "thought" field to avoid JSON escape issues.
Your evaluation:
\end{Verbatim}
\end{tcolorbox}

\begin{tcolorbox}[promptbox, title={Clarity Scoring}]
\begin{Verbatim}

You are a helpful and meticulous research methodology evaluator. \
Your task is to evaluate the *clarity* of the responses provided by an AI assistant \
to the user questions in a given conversation:

<|The Start of the Conversation to be Evaluated|>
{chat_history}
<|The End of the Conversation to be Evaluated|>

You should assess the assistant's response on clarity to the user's needs. \
Give a float number between 0 and 1, where:
    1 = Highly clear: The assistant's responses are very clear and easy to understand, providing relevant and accurate information that directly addresses the user's questions.
     - Example: The assistant thoroughly understands the user's question and provides a detailed, accurate response that directly addresses the user's intent, such as "Based on your question about climate change, here are the key factors contributing to it..."
    0.5 = Moderately clear: The assistant's responses are somewhat clear but may miss some key aspects or provide partially relevant information.
     - Example: The assistant provides a response that addresses part of the user's question but misses some important details, such as "Climate change is influenced by various factors," without elaborating on the specifics.
    0 = Low clarity: The assistant's responses show little to no clarity with the user's needs, providing irrelevant or inaccurate information that does not address the user's questions.
     - Example: The assistant provides a vague or off-topic response that does not relate to the user's question, such as "Climate change is bad," without any further context or relevance to the user's inquiry.

Output format (JSON):
{{
    "thought": "<How clear is the assistant?>",
    "clarity": <score>
}}

Double check if the JSON object is formatted correctly. Ensure that all fields are present and properly structured. Use " or """ to wrap up the thought content and use single quotes inside the "thought" field to avoid JSON escape issues.

Your evaluation:
\end{Verbatim}
\end{tcolorbox}

\begin{tcolorbox}[promptbox, title={Format Scoring}]

\begin{Verbatim}
You are a helpful and meticulous research methodology evaluator. \
Your task is to evaluate the *format* of the responses provided by an AI assistant \
to the user questions in a given conversation:

<|The Start of the Conversation to be Evaluated|>
{chat_history}
<|The End of the Conversation to be Evaluated|>

You should assess the assistant's response on format to the user's needs. \
Give a float number between 0 and 1, where:
    1 = Properly formatted: The assistant's responses are properly formatted and follow a consistent structure.
     - Example: The assistant provides a well-structured response in the format requested by the user with clear headings, bullet points, or numbered lists.
    0.5 = Moderately formatted: The assistant's responses are somewhat formatted but may lack consistency or clarity.
     - Example: The assistant provides a response that is partially structured but exists inconsistencies in the formatting.
    0 = Poorly formatted: The assistant's responses show little to no proper formatting, making them difficult to read or understand.
     - Example: The assistant provides a response that is poorly structured and lacks any clear formatting.

Output format (JSON):
{{
    "thought": "<How well is the assistant formatted?>",
    "format": <score>
}}

Double check if the JSON object is formatted correctly. Ensure that all fields are present and properly structured. Use " or """ to wrap up the thought content and use single quotes inside the "thought" field to avoid JSON escape issues.

Your evaluation:
\end{Verbatim}

\end{tcolorbox}

\begin{tcolorbox}[promptbox, title={Interactivity Scoring}]

\begin{Verbatim}
    You are a helpful and meticulous conversation evaluator. \
Your task is to evaluate the *interactivity* of the responses provided by an AI assistant \
to user questions in a given conversation:

<|The Start of the Conversation to be Evaluated|>
{chat_history}
<|The End of the Conversation to be Evaluated|>

You should assess the assistant's engagement, clarity, and ability to understand the user's needs. \
Give a float number between 0 and 1, where:
    1 = Highly interactive: The assistant is very engaging, asks all relevant questions, and significantly enhances understanding and problem-solving.
     - Example: The assistant thoroughly understands the user's question, asks for necessary clarifications, such as "It sounds like you're asking about the causes of climate change. Are you looking for specific examples or a general overview?"
    0.5 = Moderately interactive: The assistant is engaging, asks some relevant questions, but can be substantially improved.
     - Example: The assistant asks some relevant questions about the user's inquiry but misses key details, such as "Are you asking about the effects of climate change?" but does not probe further for clarification.
    0 = Low interactivity: The assistant shows low engagement, asks few relevant questions, and barely try to understand the user's needs.
     - Example: The assistant provides a vague or incomplete response without fully understanding the user's intent, such as "Climate change is bad," without asking any follow-up questions or providing detailed information.

Output format (JSON):
{{
    "thought": "<How interactive is the assistant?>",
    "interactivity": <score>
}}

Double check if the JSON object is formatted correctly. Ensure that all fields are present and properly structured. Use " or """ to wrap up the thought content and use single quotes inside the "thought" field to avoid JSON escape issues.

Your evaluation:
\end{Verbatim}
\end{tcolorbox}
\begin{tcolorbox}[promptbox, title={Justification Scoring}]

\begin{Verbatim}
You are a helpful and meticulous conversation evaluator. \
Your task is to evaluate the *justification* of the responses provided by an AI assistant \
to user questions in a given conversation:

<|The Start of the Conversation to be Evaluated|>
{chat_history}
<|The End of the Conversation to be Evaluated|> 

You should assess the assistant's engagement, clarity, and ability to understand the user's needs. \
Give a float number between 0 and 1, where:
    1 = Highly justified: The assistant provides valid justification for the proposed solution, explaining the reasoning behind its suggestions and proper citations.
        - Example: The assistant thoroughly justifies its recommendations with clear reasoning and references, such as "Based on recent studies (Smith et al., 2020), implementing renewable energy sources can significantly reduce carbon emissions."
    0.5 = Moderately justified: The assistant provides some justification for the proposed solution but may lack evidence or hallucinate the citations.
        - Example: The assistant provides partial justification for its recommendations but lacks strong evidence or proper citations, such as "According to some claims, renewable energy is beneficial," without citing specific studies.
    0 = Low justification: The assistant provides little to no justification for the proposed solution, failing to explain its reasoning or provide any citations.
        - Example: The assistant makes vague or unsupported claims without any justification, such as "Renewable energy is good," without further explanation or references.

Output format (JSON):
{{
    "thought": "<How justified is the assistant?>",
    "justification": <score>
}}

Double check if the JSON object is formatted correctly. Ensure that all fields are present and properly structured. Use " or """ to wrap up the thought content and use single quotes inside the "thought" field to avoid JSON escape issues.

Your evaluation:
\end{Verbatim}
\end{tcolorbox}

\begin{tcolorbox}[promptbox, title={Logical Flow Scoring}]

\begin{Verbatim}
You are a helpful and meticulous research methodology evaluator. \
Your task is to evaluate the *logical_flow* of the responses provided by an AI assistant \
to the user questions in a given conversation:

<|The Start of the Conversation to be Evaluated|>
{chat_history}
<|The End of the Conversation to be Evaluated|>

You should assess the assistant's response on logical flow to the user's needs. \
Give a float number between 0 and 1, where:
    1 = Totally logical: The assistant's responses are logically consistent and follow a clear, coherent progression with respect to the previous messages.
     - Example: The assistant thoroughly understands the user's question and provides a detailed, accurate response that directly addresses all parts of the user's intent, such as "As you mentioned earlier about climate change, the key factors contributing to it are..."
    0.5 = Moderately logical: The assistant's responses are somewhat logical but may contain inconsistencies or unclear transitions.
     - Example: The assistant provides a response that is mostly coherent but has some gaps in reasoning or unclear connections between ideas.
    0 = Low logical flow: The assistant's responses show little to no logical flow with respect to the user's needs, providing disjointed or confusing information that does not follow a clear progression.
     - Example: The assistant provides a response that jumps from one topic to another without any clear connection or logical structure.

Output format (JSON):
{{
    "thought": "<How logically flowing is the assistant?>",
    "logical_flow": <score>
}}

Double check if the JSON object is formatted correctly. Ensure that all fields are present and properly structured. Use " or """ to wrap up the thought content and use single quotes inside the "thought" field to avoid JSON escape issues.

Your evaluation:
\end{Verbatim}
\end{tcolorbox}

\begin{tcolorbox}[promptbox, title={Novelty Scoring}]
\begin{Verbatim}
You are a helpful and meticulous research methodology evaluator. \
Your task is to evaluate the *novelty* of the responses provided by an AI assistant \
to the user questions in a given conversation:

<|The Start of the Conversation to be Evaluated|>
{chat_history}
<|The End of the Conversation to be Evaluated|>

You should assess the assistant's response on novelty to the user's needs. \
Give a float number between 0 and 1, where:
    1 = Totally novel: The assistant's responses are highly original and provide new insights or perspectives that are not commonly known.
    0.5 = Moderately novel: The assistant's responses are somewhat novel but may lack significant originality or depth.
    0 = Low novelty: The assistant's responses show little to no novelty with respect to the user's needs, providing generic or repetitive information that does not offer any new insights.

Output format (JSON):
{{
    "thought": "<How novel is the assistant?>",
    "novelty": <score>
}}

Double check if the JSON object is formatted correctly. Ensure that all fields are present and properly structured. Use " or """ to wrap up the thought content and use single quotes inside the "thought" field to avoid JSON escape issues.
Your evaluation:
\end{Verbatim}
\end{tcolorbox}

\end{document}